\documentclass[conference,compsoc,onecolumn,11pt]{IEEEtran}
\usepackage{adjustbox}
\usepackage{amsfonts}
\usepackage{amsmath}
\usepackage{amsthm}
\usepackage{algorithm}
\usepackage{booktabs}
\usepackage{breqn}
\usepackage{bbm}
\usepackage{caption}
\usepackage{cite}
\usepackage{comment}
\usepackage{enumitem}
\usepackage[T1]{fontenc}
\usepackage{filecontents}
\usepackage{graphicx}
\usepackage{hyperref}

\usepackage{longtable}
\usepackage{lipsum}
\usepackage{makecell}
\usepackage{microtype}
\usepackage{multirow}
\usepackage{multicol}

\usepackage{pgfplots}
\pgfplotsset{compat=1.16}
\usepackage{pgfplotstable}
\usepackage{pifont}
\usepackage{stmaryrd}
\usepackage{subcaption}
\usepackage{scalerel}
\usepackage{tabu}
\usepackage{textcomp}
\usepackage{colortbl}
\usepackage{tikz}
\usepackage[framemethod=tikz]{mdframed} 
\usetikzlibrary{shapes.geometric,arrows}
\usetikzlibrary{calc}
\usepackage{thmtools}
\usepackage{thm-restate}
\usepackage{url}
\usepackage{xcolor}
\usepackage{xspace}

\usepackage{algpseudocode}

\newcommand{\subtgt}{\text{S}_f}
\newcommand{\dist}{\mathcal{D}}

\newcommand{\fm}{\mathcal{M}}

\newcommand{\rv}[1]{\mathrm{#1}}
\newcommand{\vlabel}{v}
\newcommand{\tlabel}{\Tilde{v}}
\newcommand{\logitval}[2]{\phi(#1)_{#2}}
\newcommand{\plogitval}[2]{\Tilde{\phi}(#1)_{#2}}

\newcommand{\pt}[1]{\ensuremath{p_{#1}}}
\newcommand{\mt}[1]{\ensuremath{{m_{#1}}}}
\newcommand{\world}[1]{\ensuremath{{t_#1}}}
\newcommand{\threshold}{\mathsf{{T}}}

\newtheorem{theorem}{Theorem}[section]

\newcommand{\Adv}{\ensuremath{\mathcal{A}}\xspace}

\newcommand{\D}{\ensuremath{\mathsf{D}}}

\makeatletter
\newsavebox{\@brx}
\newcommand{\llangle}[1][]{\savebox{\@brx}{\(\m@th{#1\langle}\)}%
	\mathopen{\copy\@brx\kern-0.6\wd\@brx\usebox{\@brx}}}
\newcommand{\rrangle}[1][]{\savebox{\@brx}{\(\m@th{#1\rangle}\)}%
	\mathclose{\copy\@brx\kern-0.6\wd\@brx\usebox{\@brx}}}
\makeatother

\newcommand{\Sh}{\ensuremath{\mathsf{sh}}}

\newcommand{{\piaSh}}[1]{\ensuremath{\Pi^{#1}_{\Sh}}}

\definecolor{lightmintbg}{rgb}{.53,.80,.92}

\tikzstyle{maldo} = [rectangle, minimum width=2.5cm, minimum height=0.2cm, text centered, draw=black, fill=orange!75]
\tikzstyle{hdo} = [rectangle,,minimum width=2.5cm, minimum height=0.25cm,text centered, draw=black, fill = lightmintbg!60]
\tikzstyle{malserver} = [rectangle,minimum width=0.7cm, minimum height=0.3cm, text centered, draw=black, fill=red!60]

\tikzstyle{hserver} = [rectangle,minimum width=0.7cm, minimum height=0.3cm, text centered, draw=black, fill=lightmintbg!60]
\tikzstyle{soc} = [rectangle, rounded corners,dashed,minimum width=3.6cm, minimum height=2.6cm, draw=black]

\tikzstyle{nota} = [rectangle, minimum width=8.5cm, minimum height= 0.4cm, font = \small, draw=black]

\tikzstyle{mdnota} = [rectangle, minimum width=0.4cm, minimum height=0.2cm, text centered,font = \small, draw=black, fill=orange!75]

\tikzstyle{msnota} = [rectangle,minimum width=0.4cm, minimum height=0.2cm, text centered, font = \small, draw=black, fill=red!60]

\tikzstyle{hnota} = [rectangle, minimum width=0.4cm, minimum height=0.2cm, text centered,font = \small, draw=black, fill=lightmintbg!60]
\tikzstyle{sarrow} = [ultra thin, <->,latex-latex]
\tikzstyle{darrow} = [thin,->,>=stealth]

\newlength{\maxlen}

\setlist[description]{style=unboxed,leftmargin=0cm}

\newcounter{itemcount}

\newcommand{\figlab}[1]{\label{fig:#1}}

\newenvironment{boxfig*}[2]{
	\begin{figure*}[h!]		
		\fontsize{5}{5}\selectfont
		\newcommand{\FigCaption}{#1}
		\newcommand{\FigLabel}{#2}
		\vspace{-.05cm}
		\begin{center}
			\begin{small}			 
				\begin{adjustbox}{max width=\textwidth}
					\begin{tabular}{@{}|@{~~}l@{~~}|@{}}
						\hline
						\rule[-1ex]{0pt}{1ex}\begin{minipage}[b]{.95\linewidth}
							\vspace{1ex}	
						}{%
						\end{minipage}\\
						\hline
					\end{tabular}	
				\end{adjustbox}		
			\end{small}
			\vspace{-0.1cm}
			\caption{\FigCaption}
			\figlab{\FigLabel}
		\end{center}
		\vspace{-.38cm}
	\end{figure*}
}

\newenvironment{myboxfig*}[2]{
	\begin{figure*}[!htb]		
		\fontsize{5}{5}\selectfont
		\newcommand{\FigCaption}{#1}
		\newcommand{\FigLabel}{#2}
		\vspace{-.10cm}
		\begin{center}
			\caption{\FigCaption}
			\begin{small}			 
				\begin{adjustbox}{max width=\textwidth}
					\begin{tabular}{@{}|@{~~}l@{~~}|@{}}
						\hline
						\rule[-1ex]{0pt}{1ex}\begin{minipage}[b]{.95\linewidth}
							\vspace{1ex}	
						}{%
						\end{minipage}\\
						\hline
					\end{tabular}	
				\end{adjustbox}		
			\end{small}
			\vspace{-0.25cm}
			\figlab{\FigLabel}
		\end{center}
		\vspace{-.38cm}
	\end{figure*}
}

\RequirePackage{mdframed}

\newenvironment{titlebox}[5]
{\mdfsetup{
		style=#2,
		innertopmargin=1.1\baselineskip,
		skipabove={\dimexpr0.7\baselineskip+\topskip\relax},
		skipbelow={1.5em},needspace=3\baselineskip,
		singleextra={\node[#3,right=10pt,overlay] at (P-|O){~{\sffamily\bfseries #1 }};},%
		firstextra={\node[#3,right=10pt,overlay] at (P-|O) {~{\sffamily\bfseries #1 }};},
		frametitleaboveskip=9em,
		innerrightmargin=5pt
	}
	\newcommand{\TitleCaption}{#4}
	\newcommand{\TitleLabel}{#5}
	\begin{mdframed}[font=\small]
		\setlist[itemize]{leftmargin=13pt}\setlist[enumerate]{leftmargin=13pt}\raggedright%
	}
	{\end{mdframed}
	\vspace{-1.5em}
	{\captionof{figure}{\normalfont \TitleCaption}\label{\TitleLabel}}
	\medskip
}

\tikzstyle{normal} = [thick, fill=white, text=black, draw, rounded corners, rectangle, minimum height=.7cm, inner sep=3pt]
\tikzstyle{gray} = [thick, fill=gray!90, text=white, rounded corners, rectangle, minimum height=.7cm, inner sep=3pt]

\mdfdefinestyle{commonbox}{%
	align=center, middlelinewidth=1.1pt,userdefinedwidth=\linewidth
	innerrightmargin=0pt,innerleftmargin=3pt,innertopmargin=5pt,
	splittopskip=7pt,splitbottomskip=5pt
}
\mdfdefinestyle{roundbox}{style=commonbox,roundcorner=5pt,userdefinedwidth=\linewidth}

\newenvironment{systembox}[3]
{ \begin{titlebox}{\normalfont #1}{roundbox}{normal}{#2}{#3}}
	{\end{titlebox}}

\newenvironment{gsystembox}[3]
{\vspace{\baselineskip}\begin{titlebox}{Global Functionality \normalfont #1}{roundbox}{normal}{#2}{#3}}
	{\end{titlebox}}

\newenvironment{protocolbox}[3]
{\begin{titlebox}{Protocol \normalfont #1}{commonbox}{normal}{#2}{#3}}
	{\end{titlebox}}

\newenvironment{algobox}[3]
{\begin{titlebox}{Algorithm \normalfont #1}{commonbox}{normal}{#2}{#3}}
	{\end{titlebox}}

\newenvironment{reductionbox}[3]
{\begin{titlebox}{Reduction \normalfont #1}{commonbox}{normal}{#2}{#3}}
	{\end{titlebox}}

\newenvironment{gamebox}[3]
{\begin{titlebox}{Game \normalfont #1}{commonbox}{gray}{#2}{#3}}
	{\end{titlebox}}

\newenvironment{simulatorbox}[3]
{\begin{titlebox}{Simulator \normalfont #1}{commonbox}{normal}{#2}{#3}}
	{\end{titlebox}}

\newenvironment{systembox*}[3]
{\begin{strip}
		\vspace{\baselineskip}\begin{titlebox}{Functionality \normalfont #1}{roundbox}{normal}{#2}{#3}}
		{\end{titlebox}
\end{strip}}

\newenvironment{gsystembox*}[3]
{\begin{strip}
		\vspace{\baselineskip}\begin{titlebox}{Global Functionality \normalfont #1}{roundbox}{normal}{#2}{#3}}
		{\end{titlebox}
\end{strip}}

\newenvironment{protocolbox*}[3]
{\begin{strip}
		\begin{titlebox}{Protocol \normalfont #1}{commonbox}{normal}{#2}{#3}}
		{\end{titlebox}
\end{strip}}

\newenvironment{algobox*}[3]
{\begin{strip}
		\begin{titlebox}{Algorithm \normalfont #1}{commonbox}{normal}{#2}{#3}}
		{\end{titlebox}
\end{strip}}

\newenvironment{reductionbox*}[3]
{\begin{strip}
		\begin{titlebox}{Reduction \normalfont #1}{commonbox}{normal}{#2}{#3}}
		{\end{titlebox}
\end{strip}}

\newenvironment{gamebox*}[3]
{\begin{strip}
		\begin{titlebox}{Game \normalfont #1}{commonbox}{gray}{#2}{#3}}
		{\end{titlebox}
\end{strip}}

\newenvironment{simulatorbox*}[3]
{\begin{strip}
		\begin{titlebox}{Simulator \normalfont #1}{commonbox}{normal}{#2}{#3}}
		{\end{titlebox}
\end{strip}}

\mdfdefinestyle{mycommonbox}{%
	align=center, middlelinewidth=1.1pt,userdefinedwidth=\linewidth
	innerrightmargin=0pt,innerleftmargin=2pt,innertopmargin=3pt,
	splittopskip=7pt,splitbottomskip=3pt
}
\mdfdefinestyle{myroundbox}{style=mycommonbox,roundcorner=5pt,userdefinedwidth=\linewidth}

\newenvironment{titlebox*}[5]
{\mdfsetup{
		style=#2,
		innertopmargin=0.3\baselineskip,
		skipabove={1.2em},
		skipbelow={1em},needspace=3\baselineskip,
		frametitleaboveskip=5em,
		innerrightmargin=5pt
	}
	\newcommand{\TitleCaption}{#4}
	\newcommand{\TitleLabel}{#5}
	\begin{mdframed}[font=\small]
		\setlist[itemize]{leftmargin=13pt}\setlist[enumerate]{leftmargin=13pt}\raggedright%
	}
	{\end{mdframed}
	\vspace{-1.5em}
	{\captionof{figure}{\normalfont \TitleCaption}\label{\TitleLabel}}
	\medskip
}

\newenvironment{mysystembox*}[3]
{\begin{strip}
		\vspace{\baselineskip}\begin{titlebox*}{Functionality \normalfont #1}{myroundbox}{normal}{#2}{#3}}
		{\end{titlebox*}
\end{strip}}

\newenvironment{mygsystembox*}[3]
{\begin{strip}
		\vspace{\baselineskip}\begin{titlebox*}{Global Functionality \normalfont #1}{myroundbox}{normal}{#2}{#3}}
		{\end{titlebox*}
\end{strip}}

\newenvironment{myprotocolbox*}[3]
{\begin{strip}
		\begin{titlebox*}{Protocol \normalfont #1}{mycommonbox}{normal}{#2}{#3}}
		{\end{titlebox*}
\end{strip}}

\newenvironment{myalgobox*}[3]
{\begin{strip}
		\begin{titlebox*}{Algorithm \normalfont #1}{mycommonbox}{normal}{#2}{#3}}
		{\end{titlebox*}
\end{strip}}

\newenvironment{myreductionbox*}[3]
{\begin{strip}
		\begin{titlebox*}{Reduction \normalfont #1}{mycommonbox}{normal}{#2}{#3}}
		{\end{titlebox*}
\end{strip}}

\newenvironment{mygamebox*}[3]
{\begin{strip}
		\begin{titlebox*}{Game \normalfont #1}{mycommonbox}{gray}{#2}{#3}}
		{\end{titlebox*}
\end{strip}}

\newenvironment{mysimulatorbox*}[3]
{\begin{strip}
		\begin{titlebox*}{Simulator \normalfont #1}{mycommonbox}{normal}{#2}{#3}}
		{\end{titlebox*}
\end{strip}}

\newcommand{\algoHead}[1]{\vspace{0.2em} \underline{\textbf{#1}} \vspace{0.3em}}

\makeatletter
\algnewcommand{\ExtendedState}[1]{\State
	\parbox[t]{\dimexpr\linewidth-\ALG@thistlm}{\hangindent=\algorithmicindent\strut\hangafter=3#1\strut}}
\makeatother

\algnewcommand\algorithmicinput{\textbf{Input:}}
\algnewcommand\Input{\item[\algorithmicinput]}

\algrenewcommand{\algorithmiccomment}[1]{{\color{gray}// #1}}

\newcommand{\xmath}[1]{\ensuremath{#1}\xspace}

\newcommand{\Func}[1][\relax]{\xmath{\mathcal{F}_{\textsc{#1}}}}

\newcommand{\harsh}[1]{{\color{red}{Harsh: #1}}}

\newcommand{\myparagraph}[1]{\smallskip \noindent \textbf{#1.}}
\newcommand{\revision}[1]{\textcolor{black}{#1}}

\newcommand{\system}{\text{SNAP}}

\def\extendedVersion{1}

\usepackage{enumitem}
\setlist{topsep=0pt, leftmargin=*}

\begin{document}


\title{SNAP: Efficient Extraction of Private Properties with Poisoning}

\author{\IEEEauthorblockN{Harsh Chaudhari\IEEEauthorrefmark{1},
John Abascal\IEEEauthorrefmark{1}, Alina Oprea\IEEEauthorrefmark{1},
Matthew Jagielski\IEEEauthorrefmark{2}, Florian Tram\`er\IEEEauthorrefmark{3},
Jonathan Ullman\IEEEauthorrefmark{1}}
\IEEEauthorblockA{\IEEEauthorrefmark{1}Northeastern University,
\IEEEauthorrefmark{2}Google Research,
\IEEEauthorrefmark{3}ETH Zurich}}

\maketitle

\if\extendedVersion1
\author{\IEEEauthorblockN{Harsh Chaudhari}
	\IEEEauthorblockA{Northeastern University}
	\and
	\IEEEauthorblockN{John Abascal}
	\IEEEauthorblockA{Northeastern University}
		
	\and
	\IEEEauthorblockN{Alina Oprea}
	\IEEEauthorblockA{Northeastern University}
	
	\and
	\IEEEauthorblockN{Matthew Jagielski}
	\IEEEauthorblockA{Google Research}
	
	\and
	\IEEEauthorblockN{Florian Tram\`er}
	\IEEEauthorblockA{ETH Zurich}
    
    \and
	\IEEEauthorblockN{Jonathan Ullman}
	\IEEEauthorblockA{Northeastern University}}

\fi

\begin{abstract}

Property inference attacks allow an adversary to extract global properties of the training dataset from a machine learning model. Such attacks have privacy implications for data owners  sharing their datasets to train machine learning models. Several existing approaches for property inference attacks against deep neural networks have been proposed~\cite{Ganju18,Zhang21,Chase21}, but they all rely on the attacker training a large number of shadow models, which induces a large computational overhead.
 
In this paper,  we consider the setting of property inference attacks in which the attacker can
{poison} a subset of the training dataset and query the trained target model. Motivated by our theoretical analysis of model confidences under poisoning, we design an efficient property inference attack, \system, which obtains higher attack success and requires lower amounts of poisoning than the state-of-the-art poisoning-based property inference attack by Mahloujifar et al. ~\cite{Chase21}.  For example, on the Census dataset, \system\ achieves  34\% higher success rate than \cite{Chase21} while being $\mathbf{56.5 \times}$ faster. We also extend our attack to infer whether a certain property was present at all during training and estimate the exact proportion of a property of interest efficiently. We evaluate our attack on several properties of varying proportions from four datasets and demonstrate \system's generality and effectiveness. An open-source implementation of \system\ can be found at \url{https://github.com/johnmath/snap-sp23}.
 
 \end{abstract}

\section{Introduction}



The adoption of machine learning (ML)  in a variety of critical applications  raises many privacy risks for users contributing  datasets for ML  training. In a \emph{property inference} attack~\cite{Ateniese15,Ganju18,Zhang21,Chase21} (also called \emph{distribution inference}~\cite{DistributionInference}), an adversary with query access to a trained model infers global properties of a training dataset, for example, the fraction of people belonging to a certain demographic group or with a rare disease. Some of the dataset properties leaked through these property inference attacks might reveal sensitive information an attacker can use to its advantage. As an example, a company mounting a property inference attack on an ML model released by its competitor can learn the demographic information of the competitor's clients, and adjust its targeted advertising policy for monetary gain. 


Property inference attacks proposed in the literature demonstrated that global training data properties can be inferred for deep neural networks~\cite{Ganju18,Zhang21,DistributionInference}. Recently, Mahloujifar et al.~\cite{Chase21}  showed  that poisoning the training dataset of an ML model can improve the success of property inference attacks. This threat model becomes relevant in the context of collaborative machine learning, in which users contribute their datasets for training ML models and adversaries can control a fraction of the training dataset with relatively low effort.  The main limitation of existing approaches~\cite{Ganju18,Zhang21,DistributionInference}, including~\cite{Chase21}, is that their design relies on the attacker learning a meta classifier over training examples generated from hundreds and thousands of so-called \emph{shadow models}. This meta classifier technique, which has been used in other privacy attacks such as membership inference~\cite{shokri2017membership}, incurs a large computational cost and does not always lead to the optimal attack. 


In this paper, we consider the setting of property inference attacks, in which the attacker has the capability to poison a subset of the training dataset and obtains ML model confidences for selected queries. The goal of the adversary is to learn global properties of the underlying dataset used for training the model. We address the question of how to design a property inference attack that is more efficient than those from previous work, requires lower poisoning rates, and achieves higher attack success. Keeping these goals in mind, we introduce a novel property inference attack called \system\ ({\bf S}ubpopulation I{\bf N}ference {\bf A}ttack with {\bf P}oisoning) that meets all these requirements by leveraging the insight that data poisoning attacks mounted for properties of interest create a separation between the model confidences trained with different proportions of the property. Our attack design is motivated by a novel theoretical analysis of model confidences under poisoning, leading to an efficient distinguishing test based on learning the distribution of model confidences. In particular, our attack does not require training a meta classifier, but relies on a small number of shadow models (at most 4) to learn the distribution of model confidences. This offers a significant improvement in efficiency compared to prior work~\cite{Chase21}. \revision{We also extend our attack to a label-only threat model, in which the adversary only has access to the predicted labels of the target model.} We design attacks for several property inference tasks, including: (1) distinguishing between models trained on two different fractions of the target property; (2) checking a property's existence in the training set; and (3) inferring the exact size of the property used in training. 


We evaluate our attacks comprehensively on logistic regression and neural network models trained on several datasets (Adult~\cite{Dua:2019}, Census~\cite{Dua:2019}, Bank Marketing~\cite{Dua:2019}, and CelebA~\cite{CelebA}) and a large set of 18 properties which constitute different fractions of the training data. We show that our attacks require low poisoning rates to be highly effective and are extremely efficient compared to previous attacks. To distinguish between a property present in either 1\% or 3.5\% of the Census dataset, a small poisoning rate of 0.4\% suffices to reach an attack accuracy of 96\%. To check if a property is present in the training set at all we require at most 8 poisoning samples to obtain attack accuracy higher than 95\%. We compare our \system\ attack to the state-of-the-art property inference attack by Mahloujifar et al.~\cite{Chase21} and show that \system\ consistently achieves higher attack success for multiple properties on  the Census dataset, while being $56.5\times$ more efficient.
For instance, when distinguishing the proportion of Females in the dataset, our attack achieves $91\%$ accuracy, while~\cite{Chase21}  obtains  $57\%$ success at 3\% poisoning for a logistic regression model.

\myparagraph{Our Contributions} To summarize, our main contributions are as follows:

\begin{itemize}
    \item We propose an efficient property inference attack, \system, based on an effective poisoning attack and distinguishing test between ML model confidences under poisoning. Our attack strategy is motivated by our theoretical analysis of the impact of poisoning on model confidence scores. 
    
    \item We extend our attack to perform \emph{property existence} to determine if a certain property is represented \emph{anywhere} in the training data, and estimate the \emph{exact proportion} of the property of interest. 
    
    \item We evaluate our attacks on four datasets and several ML models with a large set of  properties of different sizes. We show that our attack strategy generalizes across small, medium, and large properties, and the attack success exceeds 90\% at low poisoning rates. 
    
    \item We show that our attack improves upon the state-of-the-art property inference attack~\cite{Chase21}, while being $56.5 \times$ faster.   
    
\end{itemize}

\section{Background and Related Work} 

This section includes the required background on neural networks and the related work on existing privacy, poisoning, and property inference attacks. 

\subsection{Machine Learning Background}

 

Supervised learning encompasses a range of techniques for training ML models from labeled data. 
To train a model, a training dataset $D$ including $d$-dimensional feature vectors $X \subseteq R^d$ and class labels $Y \subseteq R^m$ is needed. The training procedure typically includes an optimization algorithm such as Stochastic Gradient Descent, to learn the model $\mathcal{M}: X \rightarrow R^m$ that minimizes a loss metric. 

Neural networks for classification learn to predict the probabilities of class labels, in addition to the label itself. For multi-class classification, the output of a neural network on input $x$ is an $m$-dimensional vector $y_1,\dots,y_m$ whose entries  sum up to 1 ($\sum_{i=1}^m{y_i} = 1$). The value $y_i$ can be interpreted as the probability that the model predicts class $i$. To generate the prediction for an input sample, the neural network performs computations across multiple layers using linear matrix operations and activation functions in each layer, to finally predict the class with the largest output probability. The neuron values at the penultimate layer $z_i$ are called \emph{logits}. The output probabilities $y_i$ of a neural network are called \emph{model confidences} and are typically computed using the softmax activation function on a model's logit values: $y_i = \mathsf{softmax}(z_i)$.  The logit values of the model can be recomputed (up to an additive shift) from model confidence as $z_i = \log\left(\frac{y_i}{1-y_i}\right)$. 


\subsection{Related Work}

\myparagraph{Individual Privacy Attacks on ML}  In many settings, ML leverages  user data to train predictive models, which might introduce a number of privacy risks for these users, as documented in previous work.

The most glaring example of privacy leakage for a user is when it is possible to reconstruct their data present in a model's training set. This has been shown to be possible for statistical databases in early work~\cite{DinurN03}, in generative language models~\cite{SecretSharer,MemorizationGPT2} and in federated learning models~\cite{InformedAdv,FL_Disaggregation,FL_Privacy} in recent work.

A less glaring form of leakage is known as a membership inference attack. Here, an adversary seeks to determine whether a given sample was present in the training set of a model~\cite{Homer+08,shokri2017membership,yeom2018privacy, Sablayrolles19, Long20, Song21, Jayaraman21, Choo21}. 
The best existing membership inference attacks train multiple models to analyze the distribution of loss \cite{EnhancedMembership} or logits \cite{LiRA} with respect to the target sample. 

\myparagraph{Poisoning Attacks in ML} Poisoning attacks assume adversarial control of a fraction of the training set. The goal of the attacker is to tamper with training data to tweak the model's behavior at inference time. 
Previous work on poisoning attacks can be classified into: \emph{availability attacks} which decrease the accuracy of models on the entire test set~\cite{Biggio2012PoisoningAA, xiao2015feature, ManipulatingML}, \emph{targeted attacks} which fool the model into misclassifying a set of target samples~\cite{NKS06,koh2017understanding, suciu2018does, geiping2020witches}, and \emph{backdoor attacks} which fool the model into reacting to a specific backdoor pattern~\cite{GLDG19,CLLLS17}. Subpopulation poisoning attacks~\cite{SubpopulationPoisoning} target specific subpopulations of data distributions. Subpopulations can be constructed by matching samples on a subset of features, or by defining clusters in the representation space of the model. 


The relationship between privacy attacks and poisoning attacks has been investigated in prior work. Ma et al.~\cite{Ma19} show that differential privacy could be a defense for poisoning attacks. Poisoning attacks have been used to improve success of privacy attacks in several settings. For instance, poisoning of private models enables auditing of private machine learning to infer lower bounds on  the privacy budget~\cite{AuditingDP, nasr2021adversary}. Also, the most recent membership inference attack ~\cite{TruthSerum}, with higher success rate than previous works \cite{EnhancedMembership, LiRA}, is  constructed with the help of data poisoning.

\myparagraph{Property Inference} Property inference attacks aim to learn global information of the training data distribution from an ML model, in contrast to attacks that leak information about individuals, such as reconstruction or membership inference attacks. Introduced by Ateniese et al.~\cite{Ateniese15}, these attacks were formalized as a distinguishing game between two worlds, where different fractions, $t_0$ and $t_1$, of the sensitive data were used to train an ML model~\cite{DistributionInference}. 
Property inference attacks can either be classified as white-box attacks~\cite{Ateniese15,Ganju18,DistributionInference}, in which the adversary has knowledge of the model architecture and parameters, or black-box attacks~\cite{Zhang21,Chase21}, in which the attacker can query the trained ML model to receive either model confidences or labels. Initial property inference attacks were designed for Hidden Markov Models and Support Vector Machines~\cite{Ateniese15}, while most of the recent papers propose attacks on deep neural networks, including feed-forward neural networks~\cite{Ganju18,Zhang21,Chase21}, convolutional neural networks~\cite{DistributionInference}, federated learning models~\cite{FeatureLeakage}, generative adversarial networks (GAN)~\cite{PI_GAN}, and graph neural networks~\cite{Inference_GNN}. Mahloujifar et al.~\cite{Chase21} showed that data poisoning can help property inference attacks achieve higher success.




We address this limitation by proposing \system, a more efficient property inference attack. \system\ uses a distinguishing test designed by following rigorous theoretical analysis to achieve higher success than previous work while also being more efficient.

%


\section{Problem Statement and Threat Model}

In this section, we introduce our problem formulation of property inference attacks and discuss the considered threat model.

\myparagraph{Property Inference}
We follow the model introduced by Ateniese et al.~\cite{Ateniese15} and used in previous property inference attacks~\cite{Ganju18,Zhang21,Chase21}. Given a dataset $D$ and a trained classifier $\fm:X \rightarrow R^m$, the goal of the adversary is to learn information about a boolean property defined on the feature space of the model $f:X \rightarrow \{0,1\}$. In particular, the adversary would like to learn the fraction of training examples satisfying the target property of interest by querying the trained model in a black-box fashion. The formalization in \cite{DistributionInference} defines a privacy game in which the adversary needs to distinguish between two worlds:

\begin{itemize}
    \item {\bf World 0}: Model $\fm$ was trained with $t_0$ fraction of training samples with property $f$;
    \item {\bf World 1}: Model $\fm$ was trained with $t_1$ fraction of training samples with property $f$.

\end{itemize}

\noindent The privacy game with poisoning capabilities defined in \cite{Chase21} proceeds as follows:
\begin{itemize}
     \item[-] Challenger $C$ selects a bit $b \in \{0, 1\}$ uniformly at random and samples a clean dataset $D_c$ of size $(1-p)n$ including  fraction $t_b$ of the property.
     
     \smallskip
     \item[-] Adversary sends a poisoned dataset {\color{black} $D_p$} of size $pn$ to the challenger.
     
     \smallskip
     \item[-] Challenger trains a model $\fm_b $ on the poisoned dataset $D_c~{\color{black}\cup~D_p}$.
     
     
     
     \smallskip
     \item[-] Adversary  queries $\fm_b$ on a set of points $x_1,\ldots, x_m$ and receives $y_1 = \fm_b(x_1),\ldots,y_m = \fm_b(x_m)$.
     
     \smallskip
     \item[-] Adversary finally outputs a guess ${\color{black} b'} \in \{0,1\}$ and wins the game if $b = {\color{black} b'}$.
     
 \end{itemize}

\myparagraph{Property Size Estimation} A generalization of the property inference formulation above, considered in~\cite{Chase21, DistributionInference}, is to allow the adversary to infer the exact size of the property of interest, without prior knowledge of the possible choices of $t_{0}$ and $t_{1}$. We consider the same setting as our property inference formulation where the adversary has the ability to poison and query a black-box model to obtain output probabilities. Instead of distinguishing between two worlds, the adversary will use the black-box model's output probabilities to perform an iterative search for the true size of the property, $t^{*}$.

\myparagraph {Property Existence} Property existence attacks can be viewed as a special case of property inference  and a generalization of membership inference. In this case the smaller fraction  $t_0$ is 0, and the adversary would like to test if there are \emph{any} samples with the target property in the training set, such that $t_1>0$. 

Property existence attacks bear some resemblance to membership inference  attacks ~\cite{shokri2017membership, LiRA,EnhancedMembership, TruthSerum}. However,  membership inference attacks test if a specific sample was present in the training set, while property existence determines if any example matching a given target property is present in the training set, without requiring complete knowledge of any particular sample. As a result, membership inference attacks may not be immediately applicable to test property inference or existence.




\myparagraph{Threat Model} \label{sec: adv_cap}
We assume that the adversary can inject  a small fraction of poisoned samples into the training dataset. This could happen in collaborative learning scenarios, in which users contribute their datasets for training ML models, and adversaries can control a part of the training set with low effort.  We would like to minimize the amount of poisoning controlled by the adversary  so that the model performance at the classification task remains similar after poisoning.  The adversary can sample training examples with and without the property of interest from the distribution of training data. 
The adversary can also query the ML model trained on the poisoned dataset to get the model output probabilities or confidence scores. We assume that the adversary is aware of the training algorithm, model architecture, features, and the number of samples used for training the target model by the model owner, but has no knowledge of the trained model parameters and the training samples.



\begin{figure*}[h]{
		\centering
			
		%
		\centering
		\begin{subfigure}[b]{\textwidth}
		    \hspace{1mm}
			\includegraphics[width=0.23\textwidth]{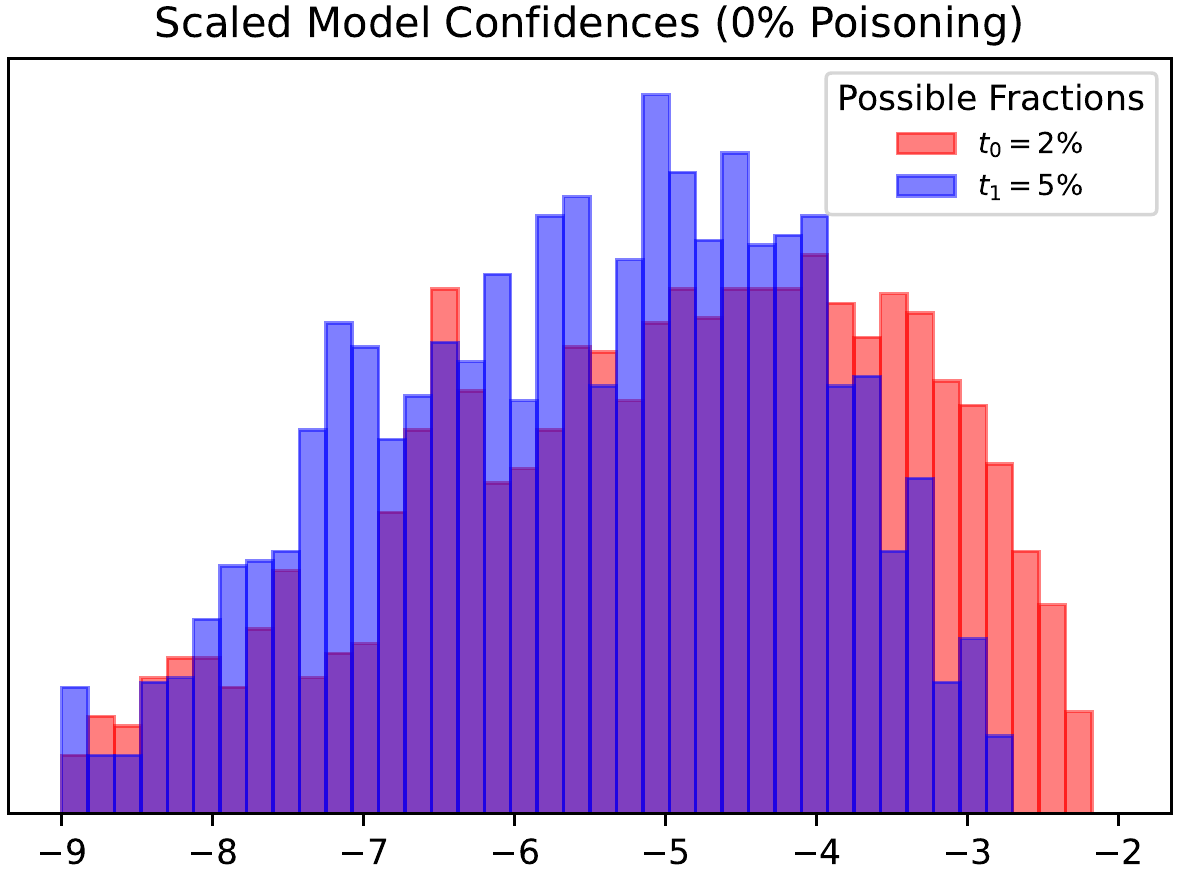}%
			\hspace{1mm}
			\includegraphics[width=0.23\textwidth]{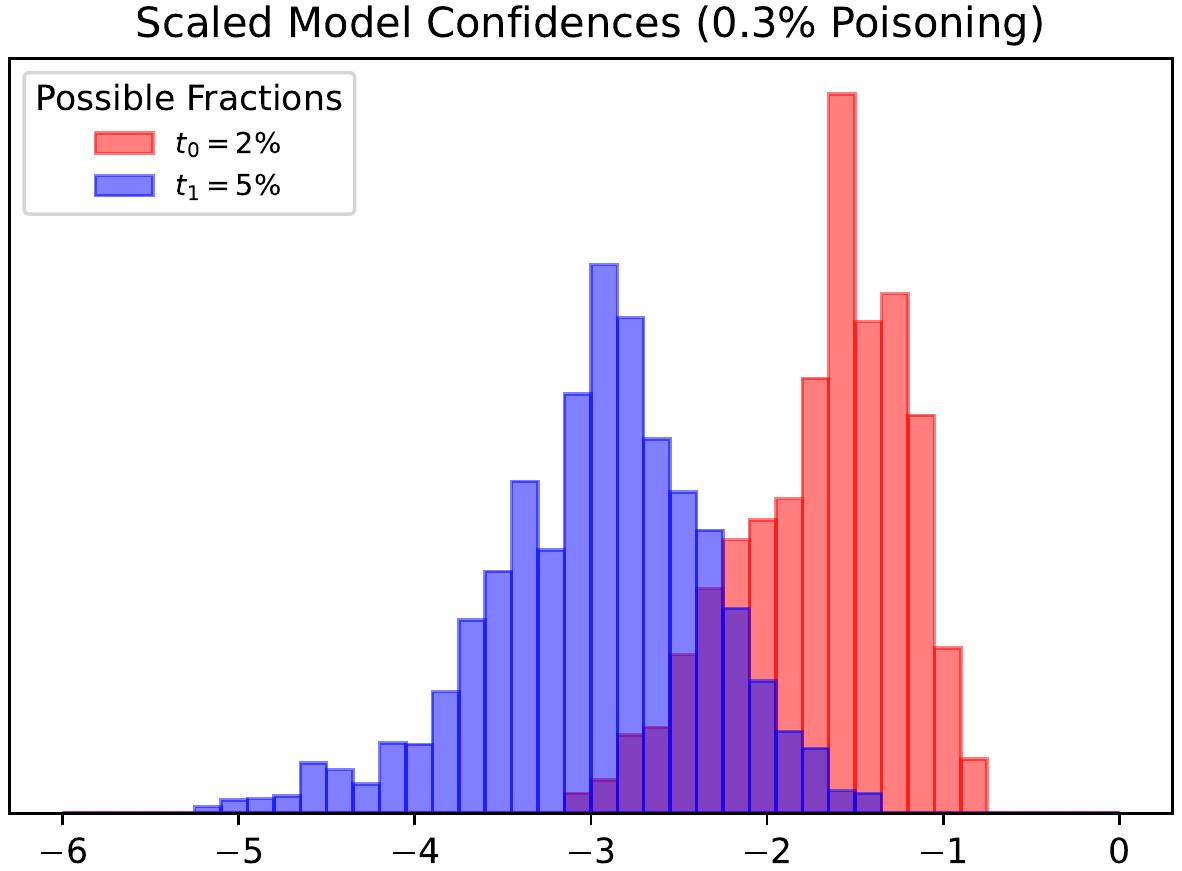}%
			\hspace{1mm}
            \includegraphics[width=0.23\textwidth]{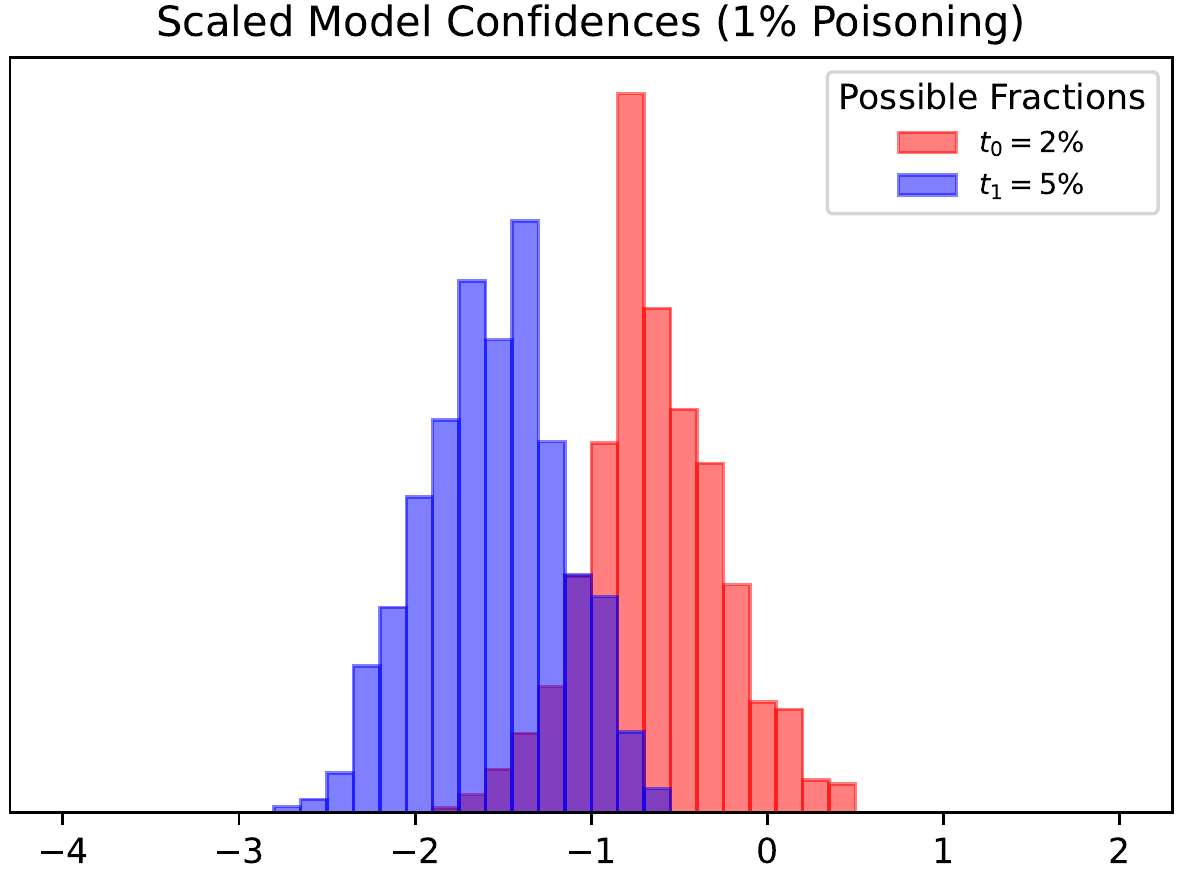}
            \hspace{1mm}
            \includegraphics[width=0.23\textwidth]{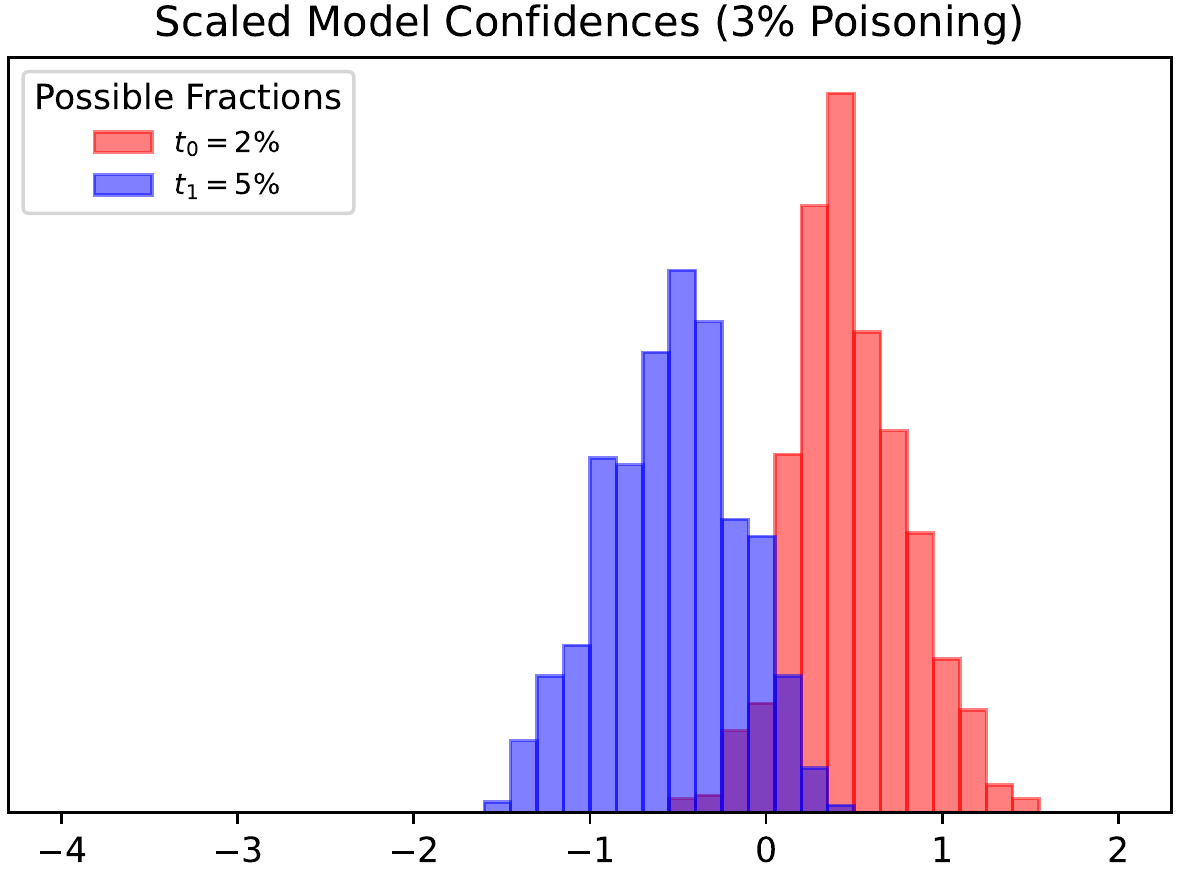}%
			\label{sfig:sub1}
		\end{subfigure}
		
	\caption{Effect of poisoning on the distribution of logit values for a given target property ``Gender = Female; Occupation = Sales'' in the Adult dataset. With increased poisoning rate, the separation between the logit distribution in the two worlds increases and the logit variance decreases. }
	
	\label{fig:LogitDist}
}
\end{figure*}

\section{Methodology}

We start by providing a brief overview of our \system\ attack in Section~\ref{sec:overview}, after which we give  the attack details  in Section~\ref{sec:attack}, introduce our theoretical analysis in Section~\ref{sec:analysis}, and finally present various extensions in Section~\ref{sec:ext}.

\subsection{\system\ Attack Overview}
\label{sec:overview}

Given the problem statement described in our previous section, there are several existing approaches in the literature for constructing property inference attacks. Recent approaches~\cite{Ganju18,Zhang21,Chase21} are based on a \emph{meta classifier}, a machine learning model trained by the attacker to distinguish the two worlds (i.e., fractions $t_0$ and $t_1$ of the target property in the training set). To generate training examples for a meta classifier, the attacker trains multiple shadow models (on the order of hundreds and thousands) for each of the worlds. These approaches draw inspiration from the literature on membership inference attacks, which uses meta classifiers and shadow models~\cite{shokri2017membership}. The main differences between existing property inference attacks are how they train the shadow models and generate training samples for the meta classifier.  In white-box settings~\cite{Ganju18} neuron values of each shadow model are sorted to generate feature vectors for the meta classifier. In black-box settings, the training examples of meta classifiers represent either model confidences~\cite{Zhang21} or labels~\cite{Chase21} from a set of queries.

In our setting, we consider a similar setup to~\cite{Chase21}, in which the attacker has black-box query access to the target model and mounts a data poisoning attack to increase the success of property inference. Our main goal is to reduce the computational complexity of property inference attacks, and achieve higher attack success at lower poisoning rates than previous work. To achieve these ambitious goals, we start by making the fundamental observation that poisoning samples with the target property impacts the two worlds differently. Using this fact, we can build an effective distinguishing test without training a meta classifier for strategically chosen poisoning rates. For instance, if we poison with a rate close to the smaller fraction $t_0$, we can change the prediction of the classifier on most of the points in World 0. The impact will be much smaller in World 1 given that $t_1 > t_0$. The difference between the poisoning success on the two worlds increases as the gap between $t_0$ and $t_1$ gets larger. In essence, we can mount a subpopulation poisoning attack~\cite{SubpopulationPoisoning} on the small world (World 0), if we treat the  target property of interest as a subpopulation. Subpopulation attacks are effective at low poisoning rates and generalize to poison the predictions of new points from the same subpopulation. This is important, as we can obtain query points for the distinguishing test by selecting points uniformly at random from the subpopulation and testing if they are misclassified.

A critical missing component of our attack is performing the distinguishing test between the poisoned models in the two worlds efficiently. Towards this, we first analyze the behavior of the logit values of samples with the property, computed by querying the poisoned models with different fractions of the property. Figure~\ref{fig:LogitDist} shows  the distributions of the logit values for two fractions  at different poisoning rates (for a property on the Adult dataset).  We observe that the logit distribution under poisoning approximately follows  a Gaussian distribution. Moreover, as the poisoning rate increases, the variance of the logit distribution decreases, leading to  a higher separation between the distributions. As a result, we  design a distinguishing test by fitting a pair of Gaussians to the two logit distributions and subsequently compute a threshold that minimizes the overlap between the two Gaussians. As we obtain a large number of samples with the property from each trained model, and the variance of the logit distribution is low, we need to train a small number of shadow models (at most 4) to estimate the mean and standard deviation of the logit distribution accurately. This leads to an exponential reduction in the number of shadow models compared to previous property inference attacks~\cite{Ganju18,Zhang21,Chase21}. Previous work modeling logit distributions for membership inference~\cite{LiRA,TruthSerum} used the logit of a single sample per model, which still required training hundreds of shadow models to estimate the logit distribution parameters. We obtain significant savings as we  model logits of all samples with the property, obtained from a small number of shadow models.   In Section \ref{sec:analysis}, we  provide theoretical analysis on the  logit distribution under poisoning, which  allows us to configure our attack effectively.

\subsection{SNAP Attack Details}
\label{sec:attack}

Our attack starts with the data poisoning step, in which poisoned samples are generated from the target property with the victim label of the attacker's choice. In the next stage, the attacker performs the model confidence learning  offline by training a small number of shadow models for each of the two worlds. The attacker learns the parameters of the Gaussian distribution of model logits for the two worlds and a separation threshold. The last stage of the attack involves the distinguishing test once  the model owner trains the target model on the poisoned set. 
Algorithm \ref{alg:adv_steps} provides an overview of the  attack strategy and we give details  below: 

\myparagraph{\revision{Data Poisoning}} \revision{Given a target property  $f$ of interest, we consider all the samples with the property as being part of a  subpopulation of the training data. The attacker chooses a victim class label $\vlabel$ that forms the majority in this target subpopulation and creates a dataset $D'$ with samples satisfying  property $f$ and having label $v$. 
The class label for each sample in $D'$ is  then changed to a target label $\tilde{v} \neq v$ of the adversary's choice to construct the poisoned dataset $D_p$ (first two steps of Algorithm \ref{alg:adv_steps}). We observe that setting the target label to the minority class in the subpopulation requires least amount of poisoning for the attack to succeed, and we thus set  $\tlabel$ to the minority class.
This form of label flipping  strategy is similar to poisoning attacks used in prior work~\cite{pmlr-v20-biggio11, SubpopulationPoisoning, Chase21}. 
The size of $D_p$ is a parameter of the attack, and  can be computed using our theoretical analysis.}

\myparagraph{Model Confidence Learning}
To distinguish models trained in the two worlds, the adversary samples points to construct datasets $D_0$ and $D_1$ with $\world{0}$ and $\world{1}$ fractions of the target property $f$, respectively. The adversary then appends the poisoned set $D_p$ to both datasets and trains $k$ shadow models per world.  The adversary constructs a query set $D_s$ 
with samples from the target property $f$ and class label $\vlabel$.
The attacker queries the $2k$ models on $D_s$, to obtain the two logit distributions.  The adversary then fits a Gaussian on both  logit distributions from the $\world{0}$ and $\world{1}$ shadow models and finally computes a threshold  $\threshold$ that minimizes the overlap between the two distributions as in Claim~\ref{prop:optThresh}. In our attack, the number of shadow models is orders of magnitude smaller than in previous work~\cite{Ganju18,Zhang21,Chase21} (i.e., at most 4) since we only use the shadow models to estimate the parameters of the target model's logit distribution (and not for meta classifier training).

\myparagraph{Distinguishing Test} In this stage, the model owner trains the target model on its dataset, which also includes the poisoning set $D_p$, and the adversary is granted black-box query access to the target model. The adversary selects a query set $D_q \subseteq D_s$, 
for querying the target  model to obtain model confidences and compute the corresponding logit values. We expect the target models' logit values to significantly overlap with either the distribution associated to World 0 or World 1. Our analysis of logits in Figure~\ref{fig:LogitDist} shows that the logit values for the smaller $\world{0}$ fraction increase faster than for $\world{1}$. The difference in the rate of the shift for the two distributions occurs because the poisoned dataset $D_p$ impacts models trained with fractions $\world{0}$ and $\world{1}$ differently: The model trained with the smaller fraction of target property examples is impacted more than the model trained with a larger fraction of examples with the target property. This is confirmed by our theoretical analysis in Section~\ref{sec:analysis}, where we prove that the means of the two logit distributions shift at different rates, and the separation increases with the amount of poisoning. 
Equipped with these observations, the adversary compares the logit values to the threshold $\threshold$ and outputs World 0 if the majority of the logit values are greater than the threshold. Otherwise, the adversary outputs World 1.




%

\begin{algorithm}[h]
	\begin{algorithmic}
		\smallskip
		\State {\bf Input} \begin{description}
			\item $f:$ Target property 
			\item $n:$ Number of training samples available to model owner
			\item $\world{0}, ~\world{1}:$  Fractions satisfying  property $f$ in the two worlds
			\item $k:$ Number of shadow models trained by attacker
			\item $p:$ Poisoned fraction of training set
			
		\end{description}
		
		\smallskip
       \State 1.  Sample  instances with property $f$  with  label $\vlabel$ to construct dataset $D' = \{(x_1,\vlabel),\ldots,(x_{p n},\vlabel) \}$ of size $pn$.
		
		
		\smallskip
		\State   2. Construct poisoned data  $D_p = \{(x_1,\tlabel),\ldots, (x_{p n},\tlabel)\}$ by changing victim label $\vlabel$ to   target label $\tlabel$.
		
		\smallskip
		\State  3. Construct datasets $D_0$ and $D_1$ of size $(1-p)n$  with  $\world{0}$ and $t_1$ fractions of  samples $x$ with target property $f(x)=1$. 
		
		\smallskip
		\State 4. Train $k$ shadow models $\fm_{0}^1,\ldots,\fm_{0}^k$ on dataset $D_0 \cup D_p$. Similarly, train $k$ shadow models $\fm_{1}^1,\ldots,\fm_{1}^k$ on $D_1 \cup D_p$.

		\smallskip
		\State 5. Construct dataset $D_s$ of  points $x$ with property $f(x)=1$ and label $\vlabel$.
		\smallskip
		\State 6. Query samples in $D_s$ on the $2k$ shadow models to obtain logit values and fit two  Gaussians  on the corresponding logit values. 
		
		\smallskip
		\State 7. Compute separation threshold $\threshold$ that minimizes  overlap between the Gaussians (See Claim \ref{prop:optThresh}).
		
		\smallskip
		\State 8.  Generate query set $D_q \subseteq D_s$  and query the samples against the  black-box target model. Obtain target model confidences and compute the corresponding logit values.
		
		\smallskip
		\State 9. If majority of logit values are larger than $\threshold$, output World 0; otherwise output World 1. 
		
	\end{algorithmic}
	\caption{SNAP Attack Strategy}
	\label{alg:adv_steps}
\end{algorithm}

\subsection{\system\ Attack Analysis} 
\label{sec:analysis}
We now analyze several aspects of our \system\ attack. We theoretically investigate the effect of poisoning on logit distributions, show how to compute an optimal separation threshold $\threshold$, and analyze the number of queries needed to succeed in the distinguishing test with high probability. The proofs of the claims are given in Appendix~\ref{apdx:AttackAnalysis}.


\myparagraph{Effect of Poisoning on Logit Distribution}
We use capital letters to denote sets (e.g., $X$) and calligraphic letters to denote distributions (e.g., $\dist$). We use $\dist_a \equiv \dist_b$ to denote equivalence of two distributions.
We use $(\rv{X},\rv{Y})$ to denote the joint distribution of two random variables. 
Notation $a \leftarrow \mathcal{A}$ denotes sampling $a$ from a distribution $\mathcal{A}$.

Consider the setting of a binary classifier. Let $\dist \equiv (\rv{X},\rv{Y}) $ denote the original data distribution of clean samples.
The attacker attempts to infer the prevalence of a property $f$. To generate the adversarial distribution $\dist_p$, they  choose a victim label $v$ (the majority label of samples satisfying $f$).
To generate a data point in  $\dist_p$, the attacker  samples a data point $(x,y) \leftarrow \dist$, such that $f(x) = 1$ and $y = v$. The attacker then assigns the target poisoned label $\Tilde{v}$ to this feature vector $x$. All points from $\dist_p$ thus satisfy the property $f$ (with original label $\vlabel$) and have poisoned label as $\Tilde{v}$.

For a property $f$, after poisoning with rate $p$, the resulting distribution $\Tilde{\dist}$ can be viewed as a weighted mixture of $\dist$ and $\dist_p$, i.e., $$\Tilde{\dist} = p \cdot \dist_p + (1-p) \cdot \dist$$
Let $(\Tilde{\rv{X}},\Tilde{\rv{Y}})$ be the joint distribution of samples from $\Tilde{\dist}$. Let $t = \Pr_{x \leftarrow \dist}[f(\rv{X}) = 1]$ be the probability that a sample satisfies property $f$ in the unpoisoned distribution and $\pi_v = \Pr[\rv{Y}=v | f(x)=1]$ the probability of label $v$ in the unpoisoned distribution for points with property $f$.
We now relate the logit of the poisoned model $\plogitval{x}{\tlabel}$ to the logit of the clean model $\logitval{x}{\tlabel}$ with respect to target label $\tlabel$. While each model's logits are likely a complicated function of their training data, we assume that they approximate the class probabilities present in the training data, so that the classifier learns $\fm(x)_v=\Pr[Y=v|X=x]$, where the probability is taken over the distribution $\fm$ was trained on.

\begin{restatable}{theorem}{LogitDist}
\label{thm:LogitDist}
For any sample $(x,y) \in D$, such that $f(x) = 1$ and $y = v$, a model $\fm$ which satisfies $\fm(x)_v=\Pr[Y=v|X=x]$ will have a poisoned logit value with respect to $\tlabel$ of
\begin{equation} \label{eqn:logit_eqn}
    \plogitval{x}{\tlabel} = \log{\left[\frac{p}{\pi_v(1-p)t} + e^{\logitval{x}{\tlabel}} \left( 1+\frac{p}{\pi_v(1-p)t}\right) \right]}
\end{equation}
\end{restatable}

For fixed $\pi_v$ and $t$, the poisoned logits will become further shifted as the poisoning rate $p$ increases. Smaller values of property fraction $t$ will be impacted more by a fixed amount of poisoning $p$, matching our intuition and making this an effective property inference test. Given the relation of the poisoned logit $\plogitval{x}{\tlabel}$ in terms of the clean logit $\logitval{x}{\tlabel}$ from Theorem~\ref{thm:LogitDist}, we  analyze the behavior of $\plogitval{x}{\tlabel}$ under the assumption that the  logit distribution is Gaussian, which empirically holds, as seen in Figure~\ref{fig:LogitDist}.

\begin{restatable}{theorem}{GaussianDist}\label{thm:GaussianDist}
Assume that the clean logit $\logitval{x}{\tlabel}$ for a sample $x$ follows a Gaussian distribution $N(\mu, \sigma^2)$. Then the mean and variance of the poisoned logit $\plogitval{x}{\tlabel}$ are $ \Tilde{u} = {\log{M} - \log{(\sqrt{\frac{V}{M^2}+1})}}$ and $\Tilde{\sigma}^2 = \log{\left( \frac{V}{M^2} + 1\right)}$ respectively, where values $M$ and $V$ denote the mean and variance of the log-normal random variable $e^{\plogitval{x}{\tlabel}}$.
\end{restatable}

\begin{figure}[h]{
		
		\begin{subfigure}[b]{\textwidth}
			\includegraphics[width=0.45\textwidth]{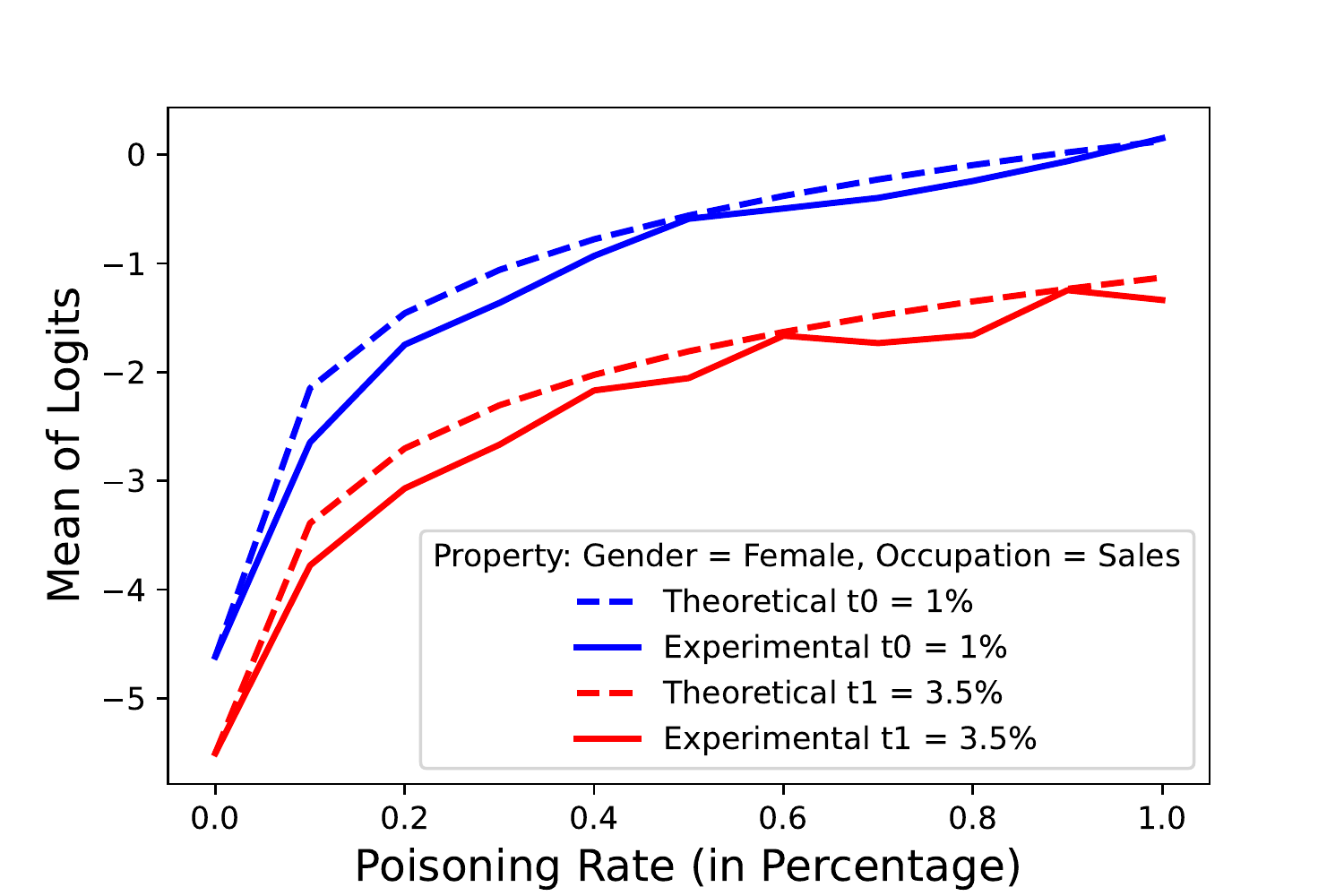}%
			\includegraphics[width=0.45\textwidth]{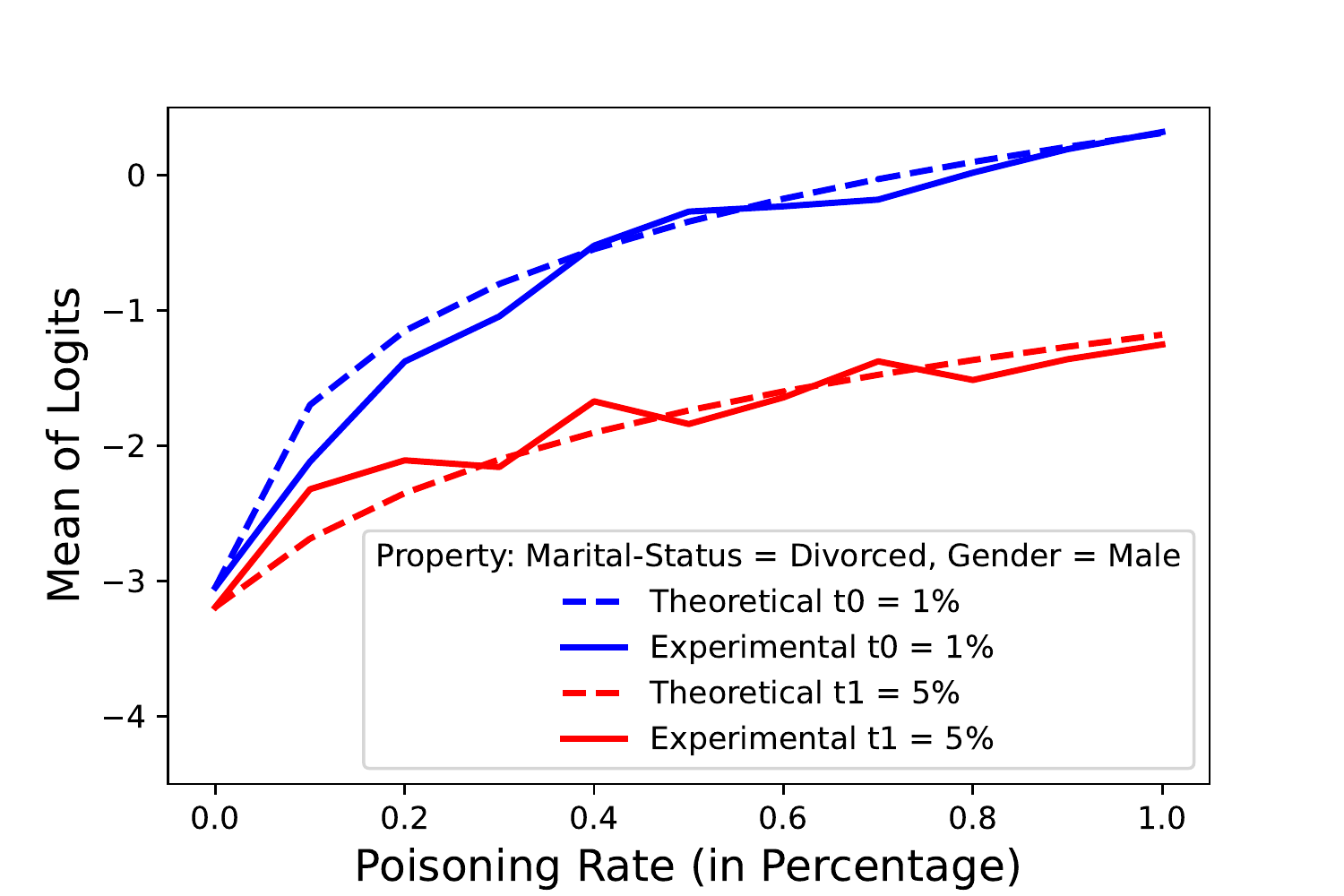}%
			\centering
			\caption{\footnotesize Shift in mean of logit distribution for models trained with $\world{0}$ and $\world{1}$ fractions of the target property by varying the poisoning rate.}
			\label{sfig:mean}
		\end{subfigure}

		\begin{subfigure}[b]{\textwidth}
		\includegraphics[width=0.45\textwidth]{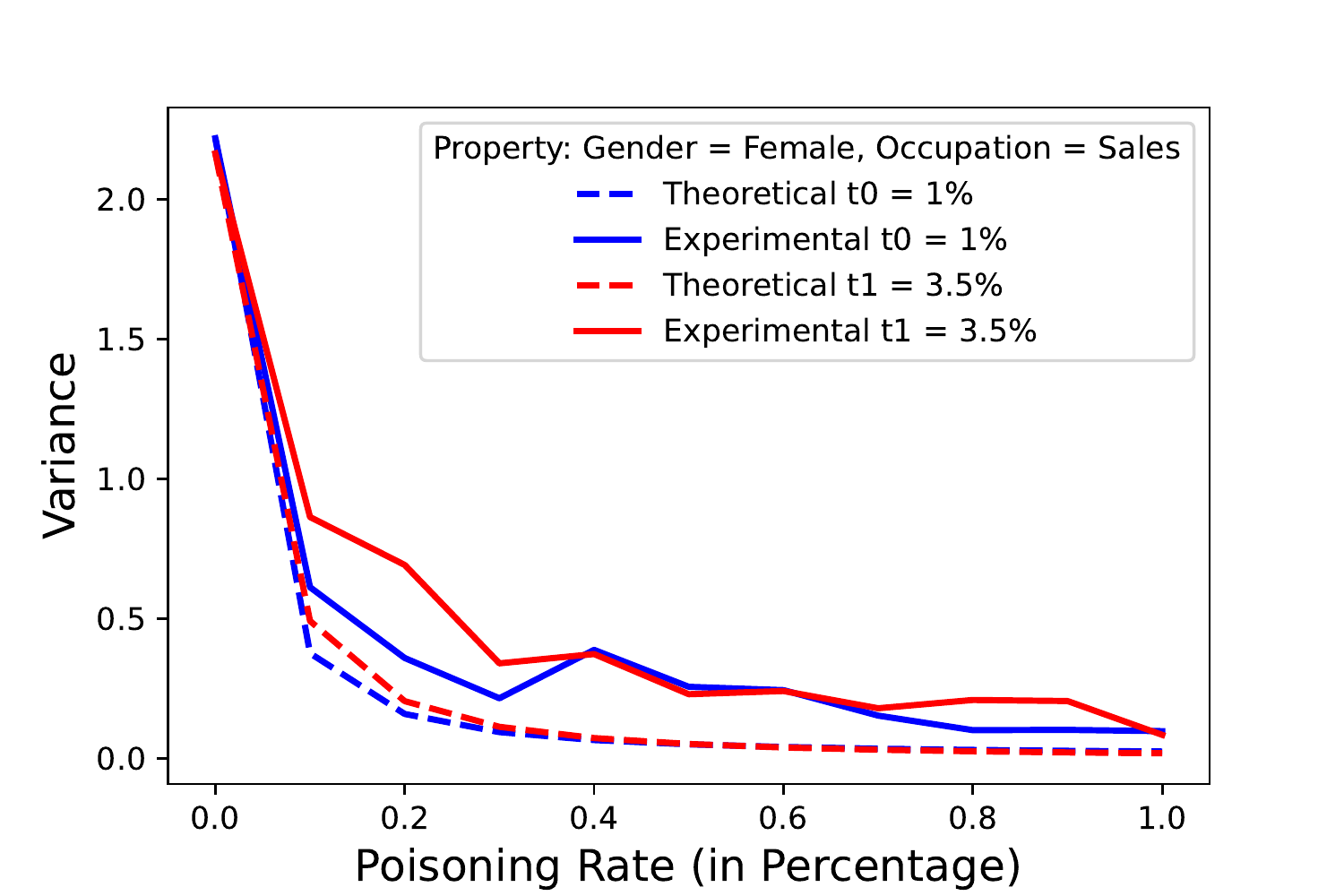}%
			\includegraphics[width=0.45\textwidth]{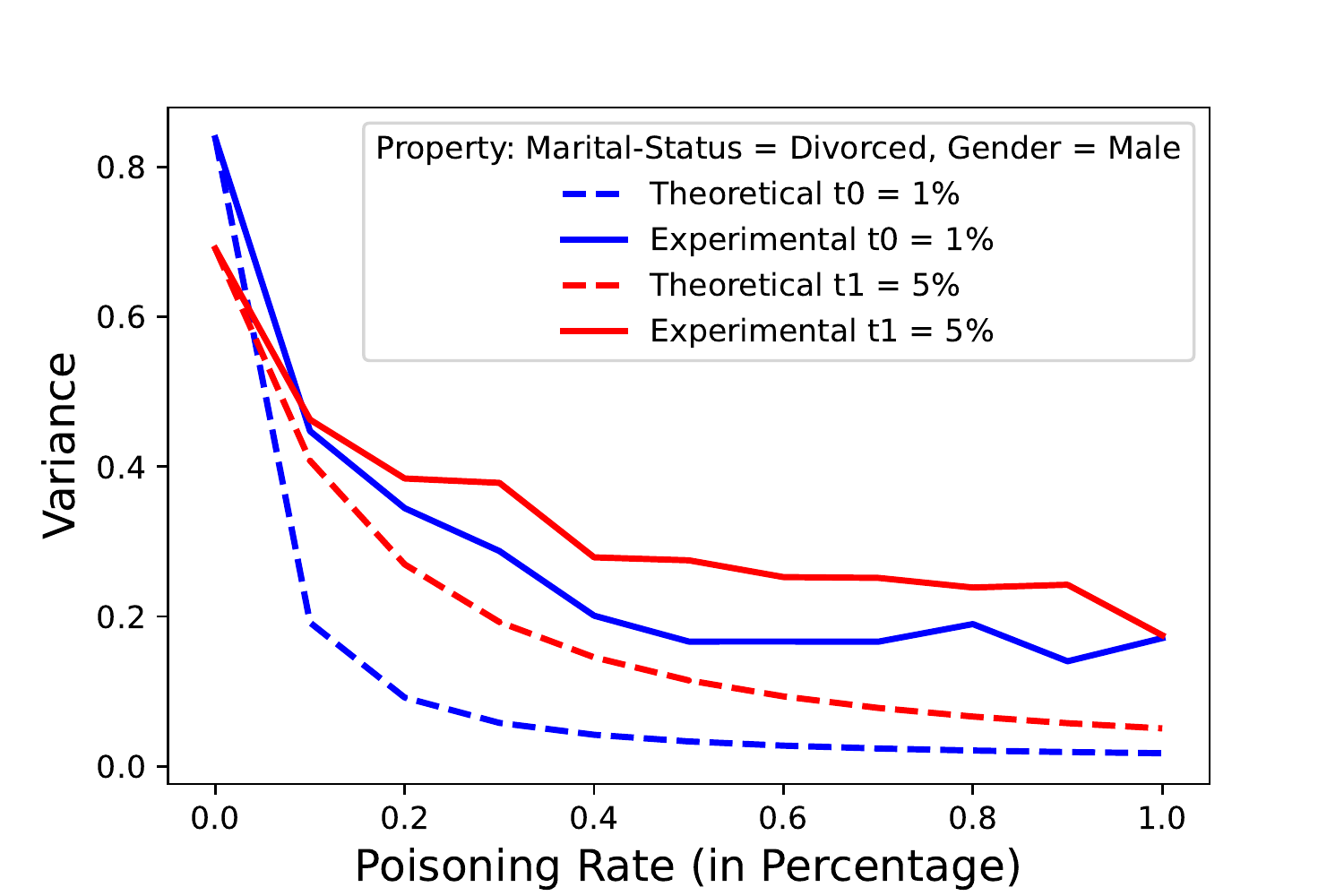}%
			\centering
			\caption{ \footnotesize Decrease in  logit distribution variance for models trained with $\world{0}$ and $\world{1}$ fractions of the target property by varying the poisoning rate.}
			\label{sfig:var}
		\end{subfigure}
		\caption{Theoretical and experimental plots on the behavior of logit values for two target properties on the Adult dataset. Experimental results confirm theoretical analysis.}
    	\label{fig:TandEResults}
}
\end{figure}

Figure~\ref{fig:TandEResults} shows how  the mean and variance of the poisoned logits vary based on our theoretical analysis compared to the experimental results. Here we train a neural network model for two properties on the Adult dataset. We observe that our analysis follows very closely to the values in our experimental results, and the relationship between the poisoned and clean logits is tight. Our analysis also confirms our observations from Figure~\ref{fig:LogitDist}: With an increase in poisoning rate,  the mean of the two distributions  shift at different rates, and the variance of both distributions shrinks, thus creating a larger separation between the two logit distributions.
\revision{ In our experiments, we observe that selecting the poisoning rate $p$ such that the theoretical variance in Theorem~\ref{thm:GaussianDist} is below a fixed threshold (e.g., $0.15$)  results in consistent high attack success larger than $90\%$ for all the properties we tested.}  





\myparagraph{Computing the Optimal Separation Threshold} Assuming the logit distribution for models trained on fractions $\world{0}$ and $\world{1}$ of the property are Gaussian, we now describe how to compute the optimal separation threshold.
%
Suppose we observe $n$ iid samples drawn from some unknown $N(\mu,\sigma^2)$. Given two hypotheses: $H_a: \mu = \mu_a$ versus $H_b: \mu = \mu_b$, we can use the Neyman-Pearson Lemma to derive an optimal test statistic and a corresponding threshold $\threshold$ that minimizes the probability of making Type-II errors (also called $\beta$) for a given significance level $\alpha$ (i.e., probability of making Type-I error). However, in our case both hypotheses $H_a$ and $H_b$ associated to fractions $\world{1}$ and $\world{0}$  are of equal importance, and, as a result, we compute a threshold that  minimizes the sum of the  probabilities of making Type-I and Type-II errors. Consequently, we  determine the optimal separation threshold as follows: 
\begin{restatable}{claim}{OptimalThreshold}\label{prop:optThresh}
Given  two Gaussian distributions $\rv{X}_0 \sim N(\mu_0,\sigma_0)$ and $\rv{X}_1\sim N(\mu_1,\sigma_1)$ such that $\mu_1 > \mu_0$ and  objective function $ J = \alpha+\beta$, where $\alpha = \Pr[\rv{X}_0>\threshold]$ and $ \beta = \Pr[\rv{X}_1<\threshold]$, the  threshold $\threshold$ that  minimizes $J$ is one of the following two values:
\begin{equation*}
	\resizebox{.47\textwidth}{!}{
	$\threshold = \frac{(\mu_0\sigma_1^2 - \mu_1\sigma_0^2) 
			\pm 2\sigma_1 \sigma_0 \sqrt{\left(\frac{\mu_1-\mu_0}{2}\right)^2 
				+(\sigma_0^2 - \sigma_1^2) \log \left(\frac{\sigma_0}{\sigma_1}\right) }}
		{\sigma_1^2 - \sigma_0^2}$
		}
\end{equation*}
\end{restatable}

In the case when the standard deviations of the two Gaussians are the same, i.e., $\sigma_0 = \sigma_1$, the separation threshold is computed as $\threshold = (\mu_0+ \mu_1)/2$. We use $\threshold = (\mu_0+ \mu_1 )/2$ as an approximation when  the standard deviations of the logit distributions are close.


\myparagraph{Number of Test Queries} 
Finally, we analyze how many samples are required in the set $D_q$ to query the target model in order for the adversary to succeed in the distinguishing test with high probability. Towards this, we provide a Chernoff bound analysis to compute the number of queries as a function of the error probabilities $\alpha$ and $\beta$. 
%
\begin{restatable}{claim}{TestQueries}\label{prop:TestQueries}
Given the probabilities $\alpha$ and $\beta$ of making Type I and Type II errors, respectively, if the adversary $\Adv$ issues $|D_q| = {\mathrm{max}}\, \left[\frac{2(2\alpha+1)\log 1/\epsilon }{(1-2\alpha)^2}, \frac{2(2\beta+1)\log 1/\epsilon }{(1-2\beta)^2} \right]$ queries, they will succeed at the distinguishing test with probability greater than $1-\max(\alpha, \beta)-\epsilon$.
\end{restatable}
The number of queries increases as $\max(\alpha, \beta)$ approaches 0.5, therefore fewer queries are needed as the distributions become  more distinguishable.

\subsection{Attack Extensions}
\label{sec:ext}

\revision{We describe several extensions of our attack to check property existence, infer properties using class labels, and estimate the size of the target property  in the training set.}

\myparagraph{Property existence} Our attack strategy remains the same  in the special case of property existence when $t_0 = 0$ and $0<t_1<1$. The goal of the adversary is to infer whether or not the target property is present in the dataset. Interestingly, our attack strategy for property existence requires much fewer poisoning samples (at most 8 samples for several tested properties) as we  demonstrate in our evaluation.

\myparagraph{\revision{\revision{Label-Only Property inference}}} \revision{
Recent work by Mahloujifar et al.~\cite{Chase21} used data poisoning to amplify property leakage under the label-only attack model in which the model returns only the predicted labels, and not the confidence scores. We construct a label-only extension of our attack for a fair comparison with~\cite{Chase21}.
Our attack uses the observation that given a sample $x$, the poisoned model will always predict the target class $\tlabel$ if the output probability on that class is greater than $0.5$. This happens only when the associated  poisoned logit $\plogitval{x}{\tlabel}$ is  greater than 0.  We use this insight to select an appropriate poisoning rate $p^*$ such that only the logits associated with the smaller world $\world{0}$ shift from negative to positive, causing the predicted label for that world to change. As an example,  in Figure \ref{fig:LogitDist}, setting $p^* = 3\%$ (the right-most figure) for target property ``Gender = Female; Occupation = Sales'' causes most of the  world  $\world{0}$ logits to become positive, while the most of the world $t_1$ logits remain negative.}

\revision{However, picking a large poisoning rate $p^*$  can be detrimental as it may cause the logits of the larger world $\world{1}$ to also become positive. We give an example in Figure \ref{fig:CompLabelOnly} (Appendix \ref{apndx:Label-Only}), where the attack accuracy of our label-only extension increases and then drops with increase in poisoning rate.
As a consequence, we propose a principled way to derive a suitable poisoning rate $p^*$ using our theoretical analysis  described in Section~\ref{sec:analysis}.
Briefly, we first show that the mean $\Tilde{\mu}$ associated to the poisoned logit $\plogitval{x}{\tlabel}$ is a strictly increasing function of the poisoning rate $p$, i.e, for a given constant $c$, there exists only one value of $p$ for which $\Tilde{\mu} = c$. We then compute an optimal $p^*$ such that $\Tilde{\mu} >0$ with respect to world $\world{0}$, while  $\Tilde{\mu}$ associated to world $\world{1}$ stays negative.  
}

\begin{figure*}[t]{
		\centering
		\begin{subfigure}[b]{\textwidth}
		    \hspace{1mm}
			\includegraphics[width=0.23\textwidth]{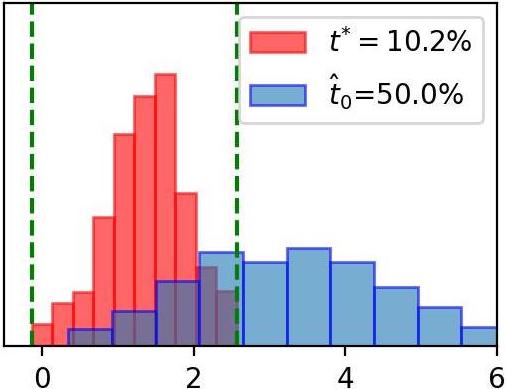}%
			\hspace{1mm}
			\includegraphics[width=0.23\textwidth]{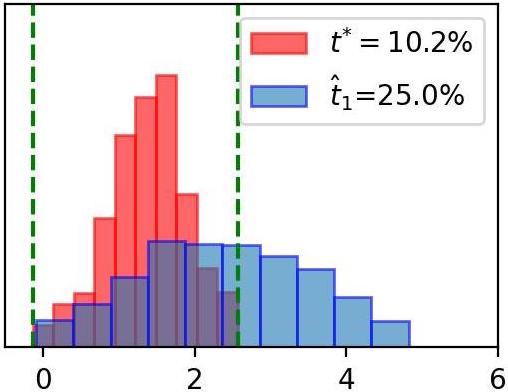}%
			\hspace{1mm}
            \includegraphics[width=0.23\textwidth]{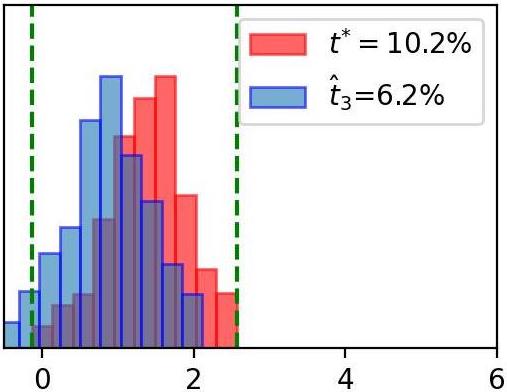}
            \hspace{1mm}
            \includegraphics[width=0.23\textwidth]{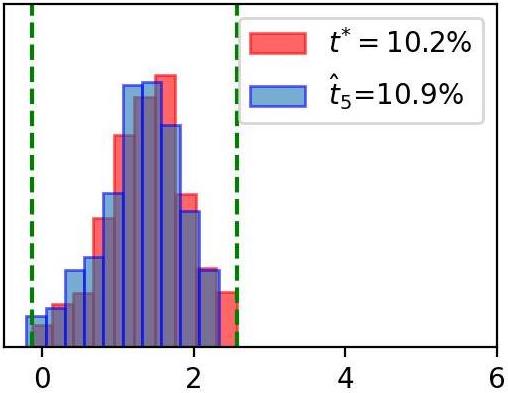}%
			\label{sfig:sub1}
		\end{subfigure}
		
	\caption{Multiple iterations of our property size estimation algorithm on Race = Black with 1\% poisoning on the Census dataset. As the algorithm progresses, the logits overlap more and the property size estimation gets closer.}
	
	\label{fig:estimation}
}
\end{figure*}

\myparagraph{Estimating property size} Given model confidences, we propose a generalization of  \system\   outlined in Algorithm 1. Instead of starting with two  guesses ($t_{0}$ and $t_{1}$) for the fraction of samples, $x$, with target property $f(x) = 1$, we estimate the \textit{true} fraction, $t^{*}$. Prior work~\cite{Chase21,DistributionInference} required thousands of shadow models to perform a distinguishing test between the worlds where the target property made up either $t_{0}$ or $t_{1}$ of the total samples. Because our distinguishing test only requires at most  4 shadow models to achieve high attack success, we can train models on datasets with several different fractions and compare their logit distributions to the target model's logit distribution when queried on $D_q$. Following this procedure introduces the same computational hurdle as observed in \cite{Chase21}'s distinguishing test: Depending on the desired precision of our estimation, we are required to train a large number of shadow models (e.g., to estimate the fraction of the target subpopulation to the nearest hundredth, we require $S=100$ shadow models to be trained and pick the one with the most similar logit distribution to the target model). This fact is made evident in the prior work, where 20,000 \cite{Chase21} and several thousand \cite{DistributionInference} shadow models are required to perform estimation with a regression meta-classifier to obtain a precision of $0.1$.

By making a key observation, we can reduce the computational complexity of the estimation: \emph{The smaller the subpopulation is in the target model's dataset, the lower the target model's prediction confidences on the subpopulation (with respect to the true labels) will be on average under data poisoning.} In other words, models trained on smaller target subpopulations will have more shifted logits, which is both empirically true and justified by Theorem~\ref{thm:LogitDist}. Additionally, more data poisoning magnifies the shifting of the logit distributions as shown in Figure~\ref{fig:LogitDist}. Using this observation, we can impose an ordering on the set of fractions we are searching over and perform \textit{binary search}, using a similarity measurement as our stopping condition. 

To perform an estimation using binary search, we  introduce the following procedure (Visualized in Figure~\ref{fig:estimation}): 1. Initialize $\hat{t}$ to 0.5; 2. Query $D_{q}$ on the poisoned target model and define $T$ as the interval from the minimum target logit value and the maximum target logit value; 3. Train shadow models on a dataset with a $\hat{t}$ fraction of the target subpopulation, and check the percentage of logits that fall within the interval, $T$; 4. If enough logits fall into the interval, stop. Else, halve the search space to include higher or lower fractions. Pick the middle of the search space to be the next $\hat{t}$. 5. Repeat until convergence. 

This method reduces the number of shadow models we need to train from $S$ to $\log_{2}(S)$. In contrast to the  thousands of shadow models required in the estimation attacks by \cite{Chase21} and \cite{DistributionInference}, our experiments required a maximum of 12 shadow models to yield estimations up to a precision of 0.001.

\section{SNAP Evaluation} 
\label{sec:Experiments}
We now evaluate the performance of our \system\ attack on four datasets: three tabular datasets (Adult, Census and Bank Marketing) and a computer vision dataset (CelebA). We select a large set of properties of different sizes (large, medium, and small) to  show the generality of our methods. We vary the attack parameters such as  poisoning rate, model complexity, size of training set, number of shadow models, and number of  queries for the distinguishing test. We also compare \system\ to prior work~\cite{Chase21} and show its improved success and performance. 

\subsection{Experimental Setup}  We first discuss the datasets and ML models used in our setup and then provide a description of the various properties considered in each dataset.

\myparagraph{Datasets and Models} We perform experiments on four datasets from different application domains (census, financial, and computer vision). The Census and CelebA datasets have been used in previous property inference papers~\cite{Ganju18,Chase21}, and we select similar properties to previous work for comparing our attack.

\begin{itemize}
    \smallskip
    \item {\em Adult: }  The UCI Adult dataset~\cite{Dua:2019} is a binary classification task with $48{,}842$ records extracted from the $1994$ Census database based on surveys conducted by the U.S. Census Bureau. Each record has $14$ demographic and employment attributes such as gender, race and marital status. The classification task is to predict whether a person's income is over $\$50{,}000$ a year.  The class label split for the dataset is 76\% and 24\% for class 0 and 1, respectively.
    We use a neural network model with two hidden layers with 32 and 16 neurons 
    after experimenting with multiple architectures, and we show later in this section results on other architectures. 
    
    \smallskip
    \item {\em Census: } The U.S. Census Income dataset \cite{Dua:2019} is a richer version of the UCI Adult dataset  containing Census data extracted from 1994 and 1995 population surveys. The dataset includes $299{,}285$ records with $41$ unique attributes.  The classification task is similar to Adult, to predict whether a person's income is over $\$50,000$ a year.  The class label split for Census dataset is 94\% and 6\% for class 0 and 1, respectively.
    We use the same two-layer neural network architecture as for Adult.
    
    \smallskip
    \item{\em Bank Marketing: }  The Bank Marketing dataset~\cite{Dua:2019} is a binary classification task with $45{,}211$ records related to marketing campaigns of a Portuguese banking institution. Each record has $16$ unique attributes such as education, occupation, month of contact, and race. The classification task is to predict if the client has subscribed a term deposit or not. The class label split for the dataset is 88\% and 12\% for class 0 and 1, respectively. We use the same two-layer neural network architecture as for Adult.
    
    \smallskip
    \item {\em CelebA: }  The CelebA dataset \cite{CelebA} contains $202{,}599$  images of celebrity faces, with each image being further annotated with a set of 40 binary attributes  such as gender, race and wearing eyeglasses. The class label split for CelebA is balanced.
    We use a ResNet-18 convolutional neural network model, where the goal of the classifier is to predict whether a person is smiling or not.  
\end{itemize}




\myparagraph{Selection of Target Properties} 
We perform our experiments on  18 different target properties across the four datasets.
Table~\ref{tab:TP}  summarizes  11 of these  properties considered for the Adult and Census datasets. The remaining 7 properties associated to the Bank Marketing and CelebA dataset are given in Table~\ref{tab:apndxTP} in Appendix~\ref{sec:AddExps}. \revision{These properties are chosen to account for a wide range of attributes and fractions across the four datasets. Two of the properties in the Census dataset and one property in CelebA were considered by  Mahloujifar et al.~\cite{Chase21}, and we include them to facilitate comparison with their approach in Section~\ref{sec:compare}.} We divide the target properties into three broad categories based on their size  relative to the size of the entire training dataset: large (above $10\%$), medium (between $1\%$ and $10\%$), and small (below $1\%$). The large and medium categories are used for property inference attacks, while the small category is used for property existence tests. Note that previous work primarily focused on large properties~\cite{Ganju18,Chase21}, but we augment the set of considered properties with medium and small ones. Even in the more challenging scenario of small and medium properties for which the separation between fractions is lower, our attacks are successful at low poisoning rates. 




{\begin{table}[h!]
		\centering 
		\begin{adjustbox}{max width=0.8\textwidth}{  
				\begin{tabular}{@{}c c c  r  r@{}}
					
					
					{Attack Type} & {Property Size} & {Dataset} & \multicolumn{1}{c}{Target Properties} & Distinguishing Test\\ \midrule
					
					\multirow{10}{*}{\makecell{Property \\Inference}} & \multirow{5}{*}{Large}  & \multirow{2}{*}{Adult} & Workclass = Private & $20\%$ vs $40\%$ \\
					
					                                    &                         &                        &  Race = White; Gender = Male & $15\%$ vs $30\%$ \\
                    \cmidrule{3-5}
                    
                                                        &                         & \multirow{2}{*}{Census} & Race = Black & $10\%$ vs $25\%$\\
                                                        &                         &                         & Gender = Female & $30\%$ vs $50\%$\\

                    \cmidrule{2-5}
                                                        & \multirow{4}{*}{Medium} & \multirow{2}{*}{Adult}   & Gender = Female; Occupation = Sales & $1\%$ vs $3.5\%$ \\
                                                        
                                                        &                        &                          & Marital-Status = Divorced; Gender = Male & $1\%$ vs $5\%$ \\ 
                    \cmidrule{3-5}
                                                        &                        & \multirow{2}{*}{Census}  & Education = Bachelors & $2\%$ vs $8\%$ \\
                                                        
                                                        &                        &                          & Industry  = Construction  & $2\%$ vs $7\%$\\
                                                        
                    \cmidrule{1-5}
                    
                    \multirow{3}{*}{\makecell{Property\\ Existence}} & \multirow{4}{*}{Small} & \multirow{2}{*}{Adult} & Native Country = Germany & $0\%$ vs $0.10\%$ \\
					                                    &                             &                        & Occupation = Protective Services & $0\%$ vs $0.05\%$ \\
                    \cmidrule{3-5}
					                                    &                             & \multirow{1}{*}{Census} & Hispanic-Origin = Cuban & $0\%$ vs $0.20\%$\\    

				\end{tabular}
			}
		\end{adjustbox}
		\caption{Target properties considered in the Adult and Census datasets. The attacker's objective is to distinguish between the two percentages  of the target property shown in the last column. } 
		\label{tab:TP}
	\end{table}
}

\myparagraph{Target and Shadow Model Training} To create training datasets for the attacker and the model owner we partition the original training set equally between the two such that the two subsets are disjoint. The attacker trains shadow models for each  fraction $t_0$ or $t_1$ of the  property of interest. The default value for the number of shadow model is 4 for each fraction $t_0$ and $t_1$, which is sufficient to learn the distribution of model logits. We vary the number of shadow models later in the section.
The remaining samples with the property $f$ and label $v$ not used for shadow model training will be part of the attacker query set $D_s$.

\myparagraph{Test Query Set} 
The attacker  requires  black-box access to the target model and obtains  the confidence scores of the  model on a set of queries,  denoted by $D_q$, which is a subset of $D_s$. Note that each sample $x$ in $D_q$  needs to satisfy the property $f(x) = 1$ and its corresponding label $\vlabel$.  \revision{ For all the target properties, we set our query set size  to 1000 samples based on our conservative analysis described in Claim \ref{prop:TestQueries} with respect to a large property and a small poisoning rate $p= 1\%$. We also vary this parameter later in Section \ref{sec:expPropInf}.}


\begin{figure*}[t]{
        \centering
        
		\begin{subfigure}[b]{0.48\textwidth}
			\includegraphics[width= \textwidth]{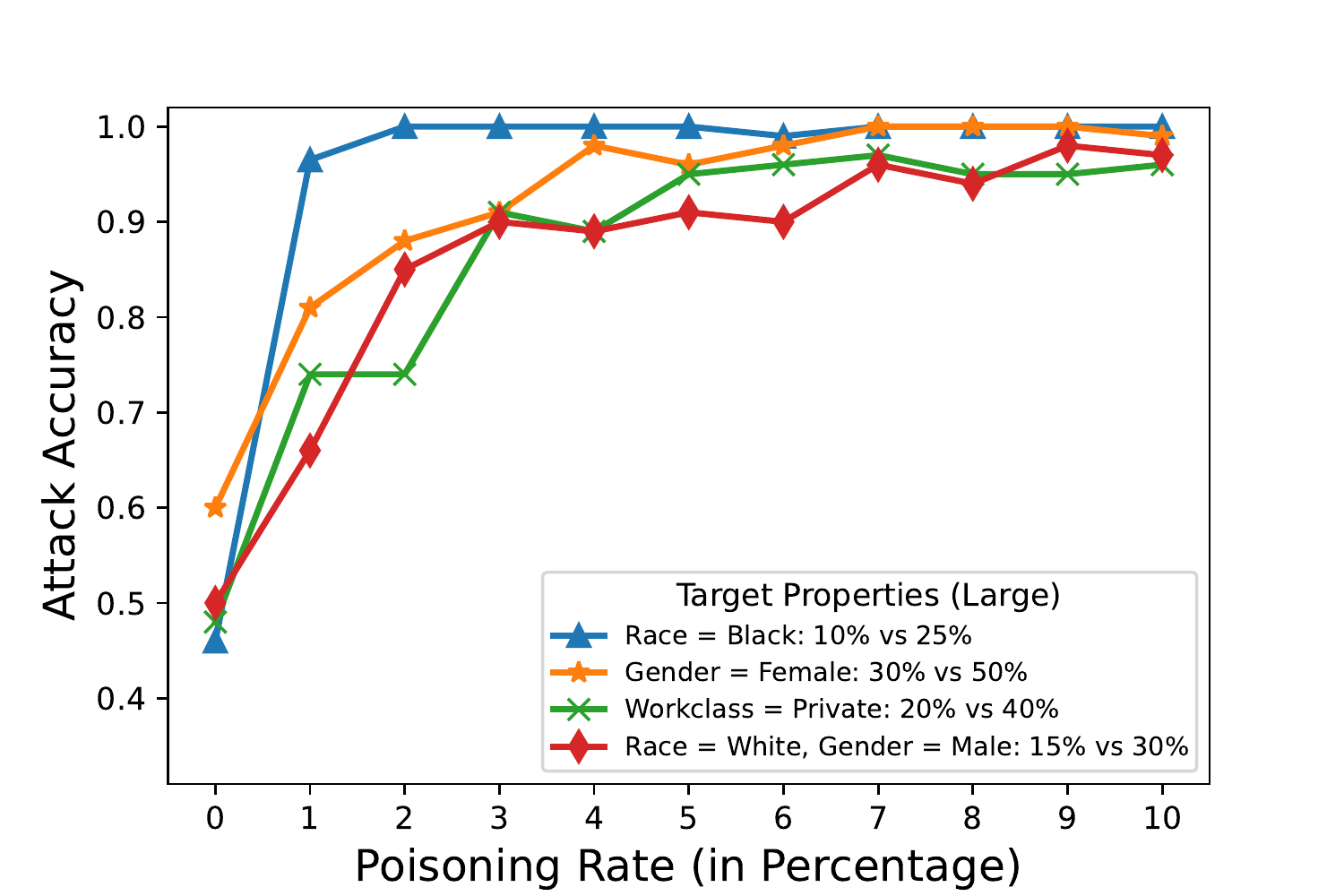}%
			
			\caption{\centering 
			Attack accuracy increases with the amount  of poisoning, and is above  90\% at 5\% poisoning for large properties. }
			\label{fig:PRLP}
		\end{subfigure}
    	\begin{subfigure}[b]{0.48\textwidth}
			\includegraphics[width=\textwidth]{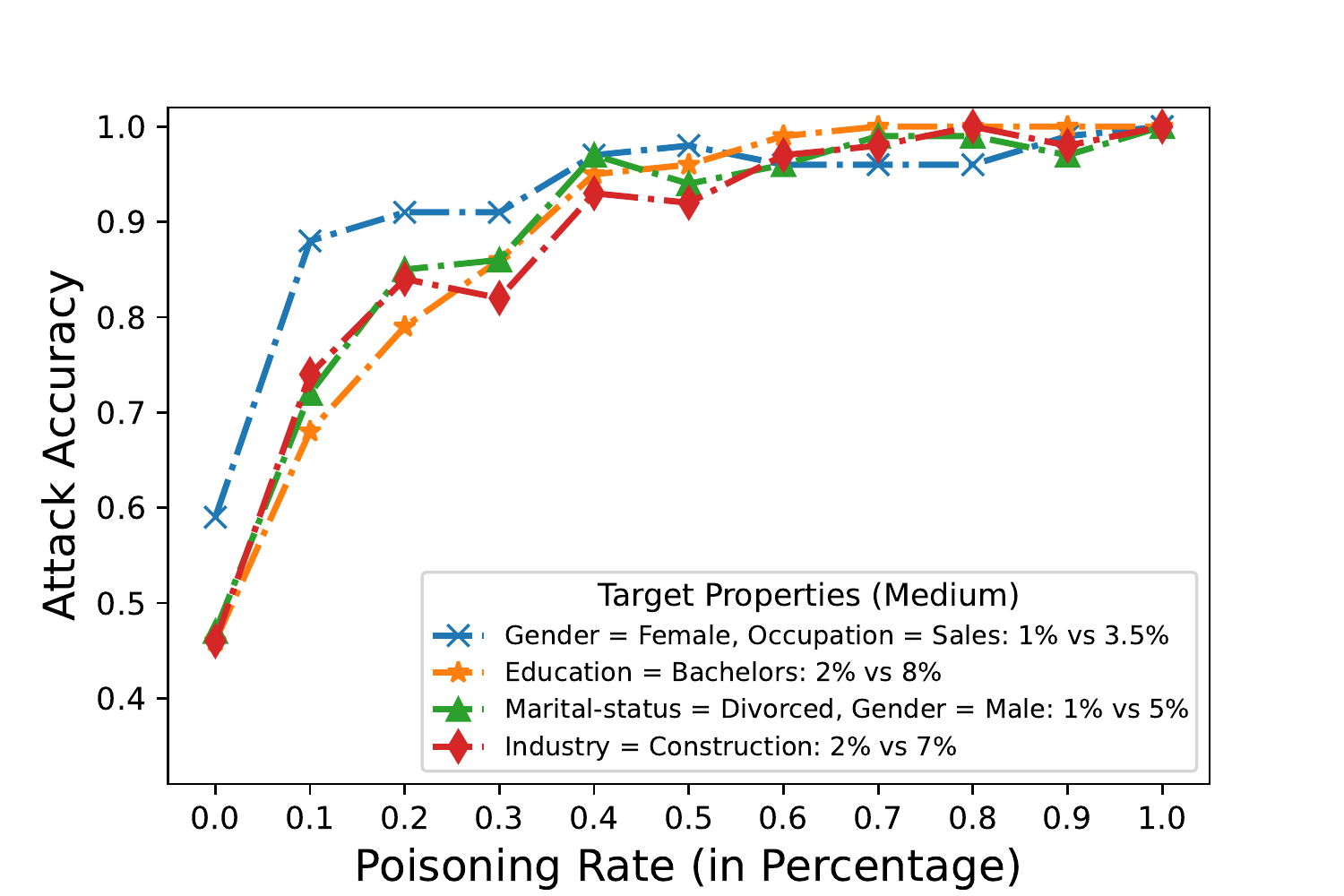}%
			\caption{ \centering
			 Medium properties require less poisoning, with attack accuracy above  90\%  at  0.6\% poisoning.}\label{fig:PRSP}
		\end{subfigure}
		
		\begin{subfigure}[b]{0.48\textwidth}
			\includegraphics[width= \textwidth]{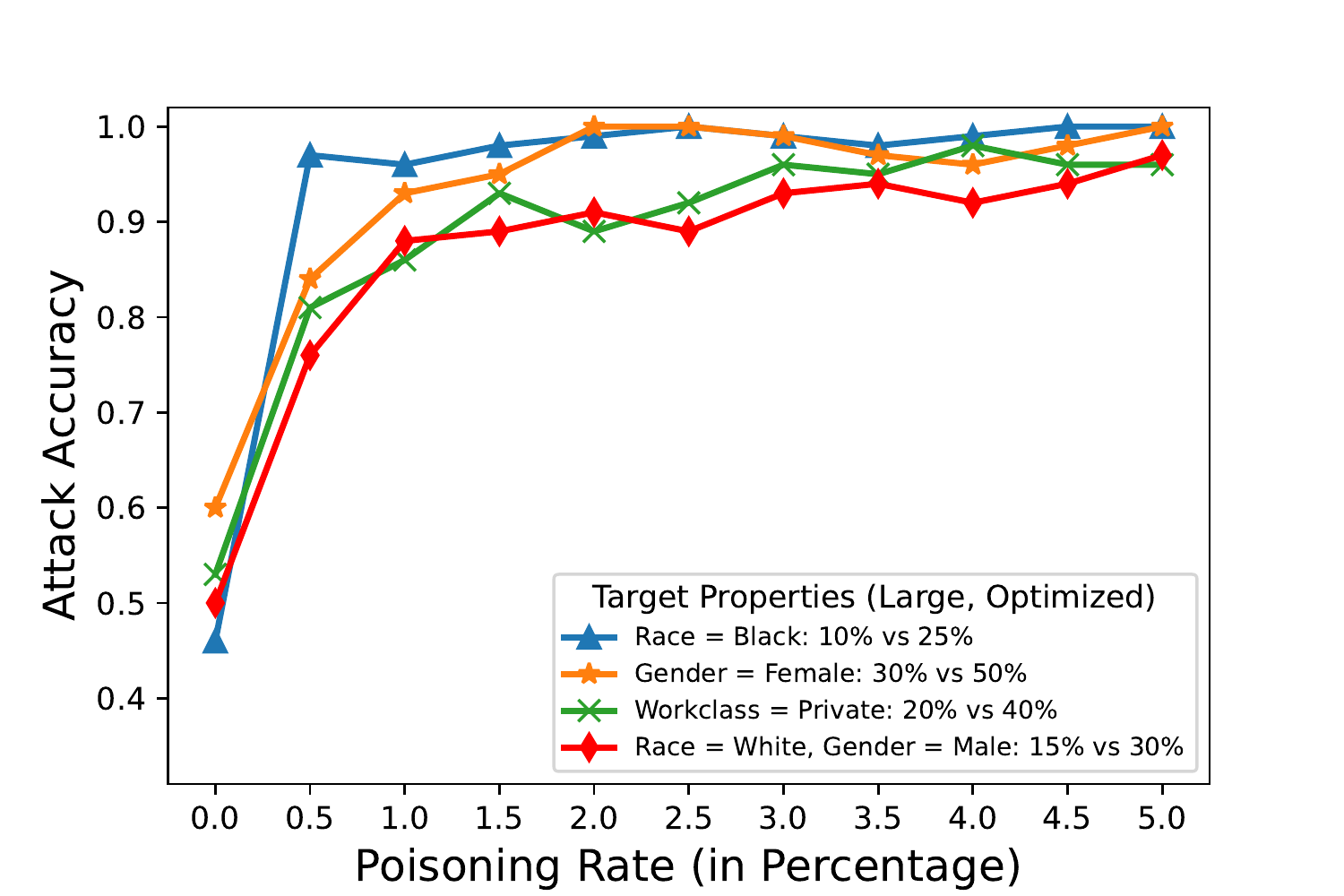}%
			
			\caption{\centering 
			  Optimized attack  requires  less poisoning for large properties, reaching 90\% accuracy at 1.5\% poisoning.}
			\label{fig:PRLPOPT}
		\end{subfigure}
	  \begin{subfigure}[b]{0.48\textwidth}
			\includegraphics[width=\textwidth]{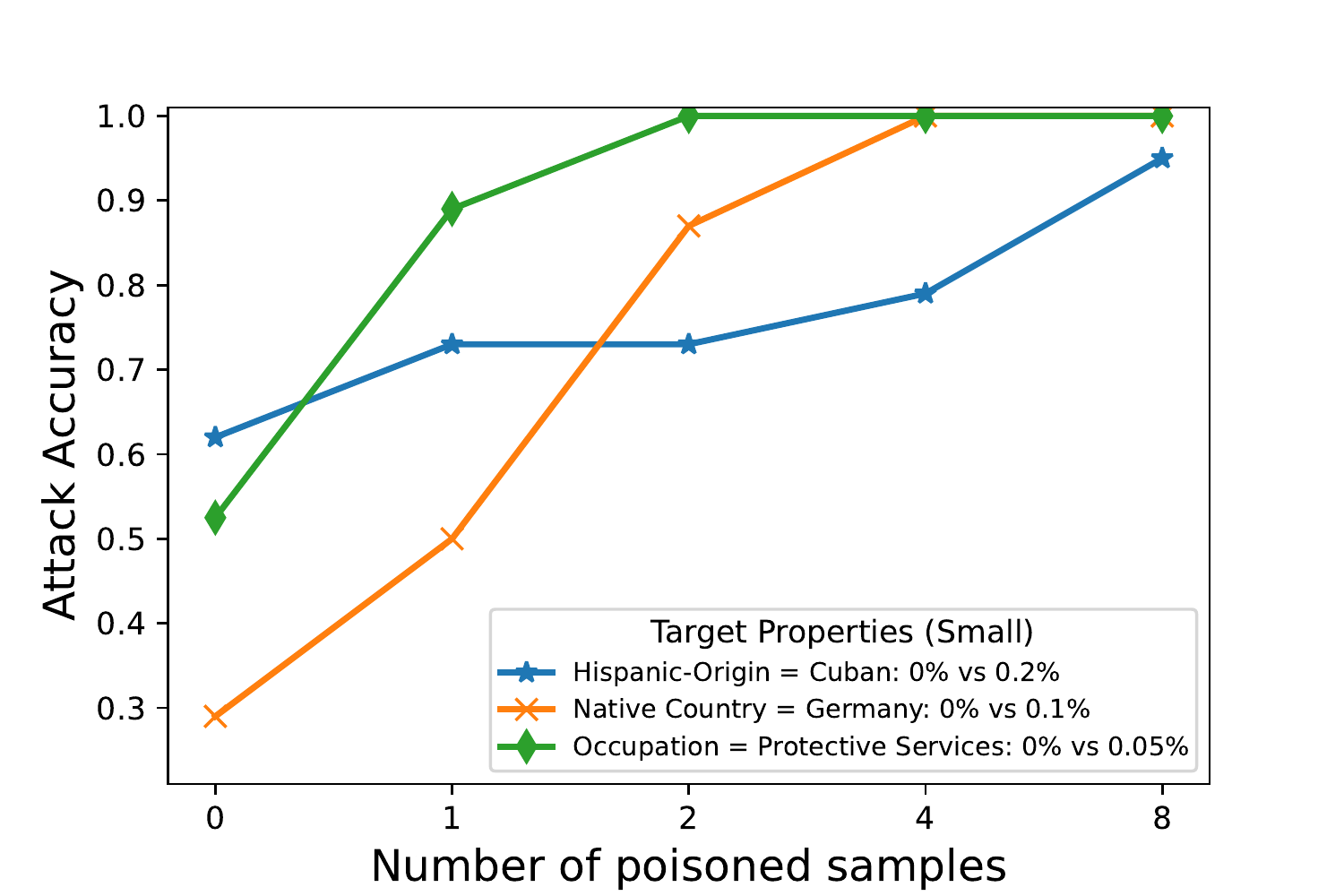}%
	        \caption{\centering 
	        Property existence attack accuracy is above 95\% with only $8$ poisoned samples.}\label{fig:PropExst}
		\end{subfigure}
		
		\caption{Attack accuracy for large, medium and small target properties on Adult and Census dataset. }
}
\end{figure*}

\myparagraph{Success Metric} 
Similar to previous works \cite{Ganju18, Chase21, Zhang21}, the attacker's success is computed in terms of accuracy of correctly distinguishing which fraction of the target property the ML model was trained on. We repeat all of our experiments $5$ times. In each of the $5$ trials, we train $10$ target models per fraction and query them on $10$ different test query sets, giving us a total of $200$ observations per trial. As a result, the reported attack accuracy is averaged over $1000$ observations.


\subsection{Property Inference Evaluation} \label{sec:expPropInf}
We first evaluate the performance of our \system\ attack depending on the amount of poisoning while using the default parameters described earlier. We then evaluate how  different parameters such as the number of shadow models, complexity of model architecture, training set size and the number of test queries 
impact the attack accuracy. To understand the impact of each parameter, we vary one parameter at a time while fixing the rest. 

\myparagraph{Amount of Poisoning}
We analyze the accuracy of our property inference attack on the target properties as we vary the poisoning rate. Figures~\ref{fig:PRLP}  and~\ref{fig:PRSP} provide results of attack accuracy for large and medium target properties, respectively. The attack accuracy for all the target properties is  low when there is no poisoning (close to the 50\% random guessing probability of the distinguishing test).  As we increase the poisoning rate, the attack accuracy improves dramatically for all properties. For large target properties, as shown in Figure \ref{fig:PRLP}, the attack accuracy reaches close to $100\%$  (perfect success) as we approach  $10\%$ poisoning, but it is above 90\% at 5\% poisoning rate. 


Previous works in property inference~\cite{Ganju18, Zhang21, Chase21}  primarily attacked large target properties and did not consider medium and small properties as we do. Performing the attack on smaller properties  is more challenging as the attacker needs to distinguish between  smaller separations. As observed in Figure \ref{fig:PRSP}, our attack is successful on all 4 medium properties (with thresholds between 1\% and 8\%), and the attack accuracy exceeds $90\%$ with as little as $0.6\%$ poisoning on the Adult and Census datasets. In fact, for distinguishing between 1\% and 3.5\% of ``Gender = Female; Occupation = Sales'' on the Adult dataset, \system\ achieves 96\% success at only 0.4\% poisoning. 

Based on the observation that our strategy for medium properties achieves high attack accuracy with low poisoning rate, we are able to attack large properties more effectively. Concretely, we can target and poison a smaller sub-property within the larger target subpopulation. For instance, for the target property ``Gender = Male; Race = White'' we originally require $5\%$ poisoning to achieve  $90\%$  attack accuracy at distinguishing between $15\%$ and $30\%$ of samples with this property. With our optimized approach, by poisoning the smaller subpopulation ``Gender = Male; Race = White; Marital-Status = Never-Married'', we require only $1\%$ poisoning to reach $90\%$ attack accuracy for the same distinguishing test between $15\%$ and $30\%$ of ``Gender = Male; Race = White'' in the whole dataset. The intuition is that a separation between worlds in the larger subpopulation ``Gender = Male; Race = White'' still results in a  separation for the sub-property, particularly if the proportion of ``Marital-Status = Never-Married'' is uniform within the ``Gender = Male; Race = White'' population, which can be ensured by selecting independent features for the sub-property. 
Table \ref{tab:apndxSubP} in Appendix \ref{sec:AddExps} includes  the  sub-properties used to attack each of the large properties and Figure \ref{fig:PRLPOPT} provides the  attack accuracy for large properties with our described optimization. Our modified version is very effective as it achieves attack accuracy close to $90\%$ with only 1.5\% poisoning across all properties. We observe similar results for the Bank-Marketing dataset, included in Figure \ref{fig:apndx_BM} in Appendix \ref{sec:AddExps}. 


We measured the precision, recall and F1 score of the original models and models poisoned at different rates. For large properties, poisoning could reduce the F1 score metrics by at most 8\%, but the metrics remain similar at low poisoning rates, as required for the medium properties and our optimized attack on large properties.



	%
	

 


\myparagraph{Complexity of Models}
So far, we have fixed the model architecture to a two-layer neural network (2NN) model for Adult and Census datasets. Here, we vary the complexity of the model from one to six layer neural networks to understand its impact on the attack accuracy. Table~\ref{tab:NNarch} provides the  model architectures chosen similarly to prior work~\cite{Chase21}. 


{\begin{table}[h!]
		\centering \scriptsize
		\begin{adjustbox}{max width=0.7\textwidth}{  
				\begin{tabular}{c |c c c c c c}
					
					
					{Model Type}  & {1NN} & {2NN} & {3NN} & {4NN} & {5NN} & {6NN} \\ \midrule
					
					{Architecture}& [32] & [32, 16] & [32, 16, 8] & [32, 16, 8, 4] &[32, 16, 8, 4, 2]  & [64, 32, 16, 8, 4, 2] \\

				\end{tabular}
			}
		\end{adjustbox}
		\caption{Model architectures, each element in the list is the number of nodes in a hidden layer of a neural network.} \label{tab:NNarch}
	\end{table}
}

We evaluate our attack accuracy and F1 score of the model on ``Race = White; Gender = Male''  and ``Gender = Female; Occupation = Sales'' target properties at poisoning rates $2\%$ and $0.2\%$, respectively. We use F1 scores instead of test accuracy as the datasets have high class imbalance. 
We observe that as the complexity of the model increases  from one layer  to two layers, the ability of the model to fit the data improves, and, consequently, the attack accuracy improves. For instance, for  ``Race = White; Gender = Male'' property,  we observe the attack accuracy and F1 score improve by $10\%$ and $6\%$ respectively.
However, as the model starts overfitting, both the F1 score and the attack accuracy  drop. The attack accuracy of the 6NN model is  lower than the 2NN model by at least $17\%$ across both target properties, which shows that  overfitting  deteriorates our attack accuracy.

\begin{figure*}[h!]
 \centering
    \begin{minipage}{0.47\textwidth}
        \includegraphics[width=\textwidth]{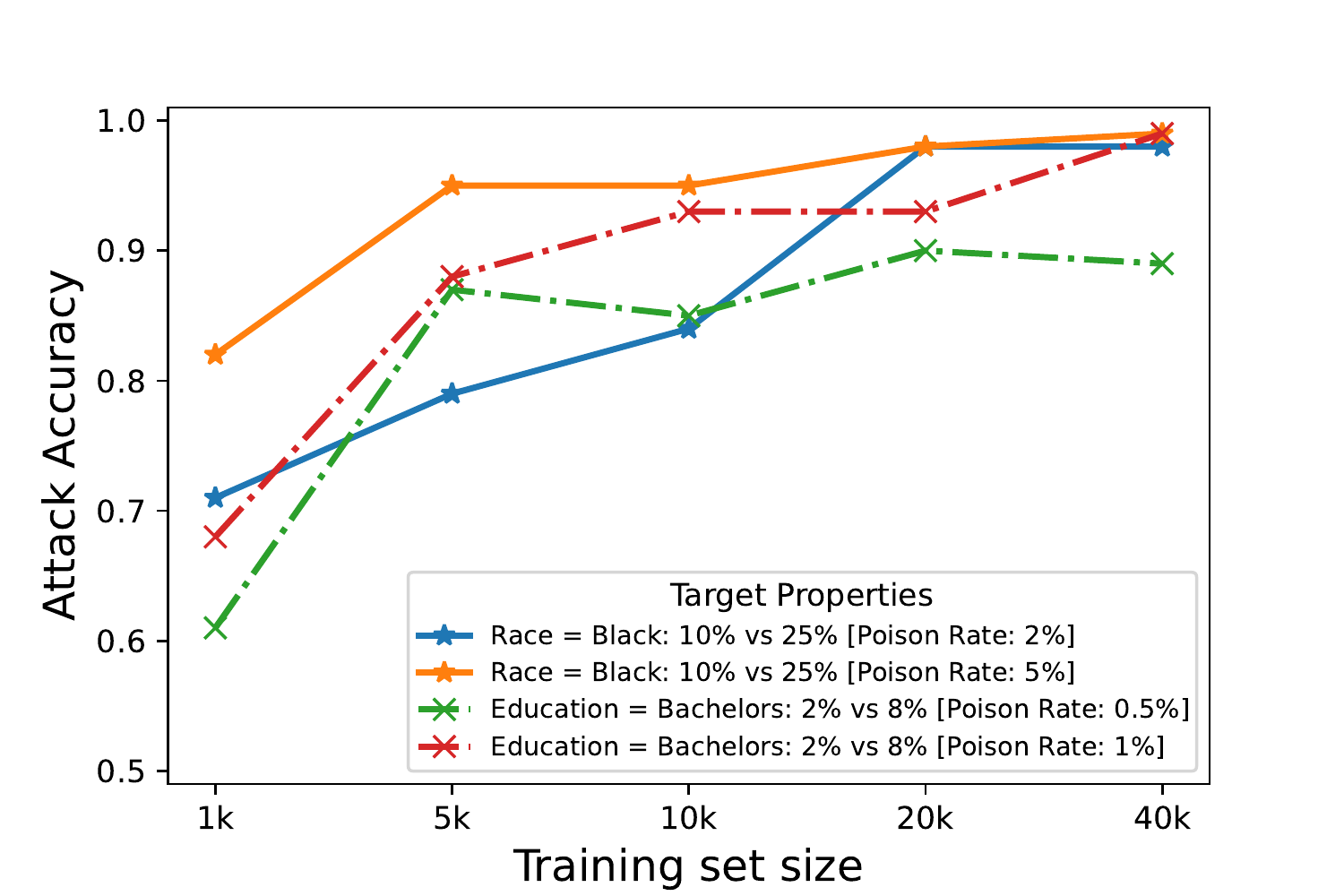}
	    \caption{\centering Attack accuracy by the training  dataset size. More training samples improve the shadow models performance, resulting in higher attack accuracy.}
	    \label{fig:TrainSS}
    \end{minipage}
      \begin{minipage}{0.47\textwidth}
        \centering
        \includegraphics[width=\textwidth]{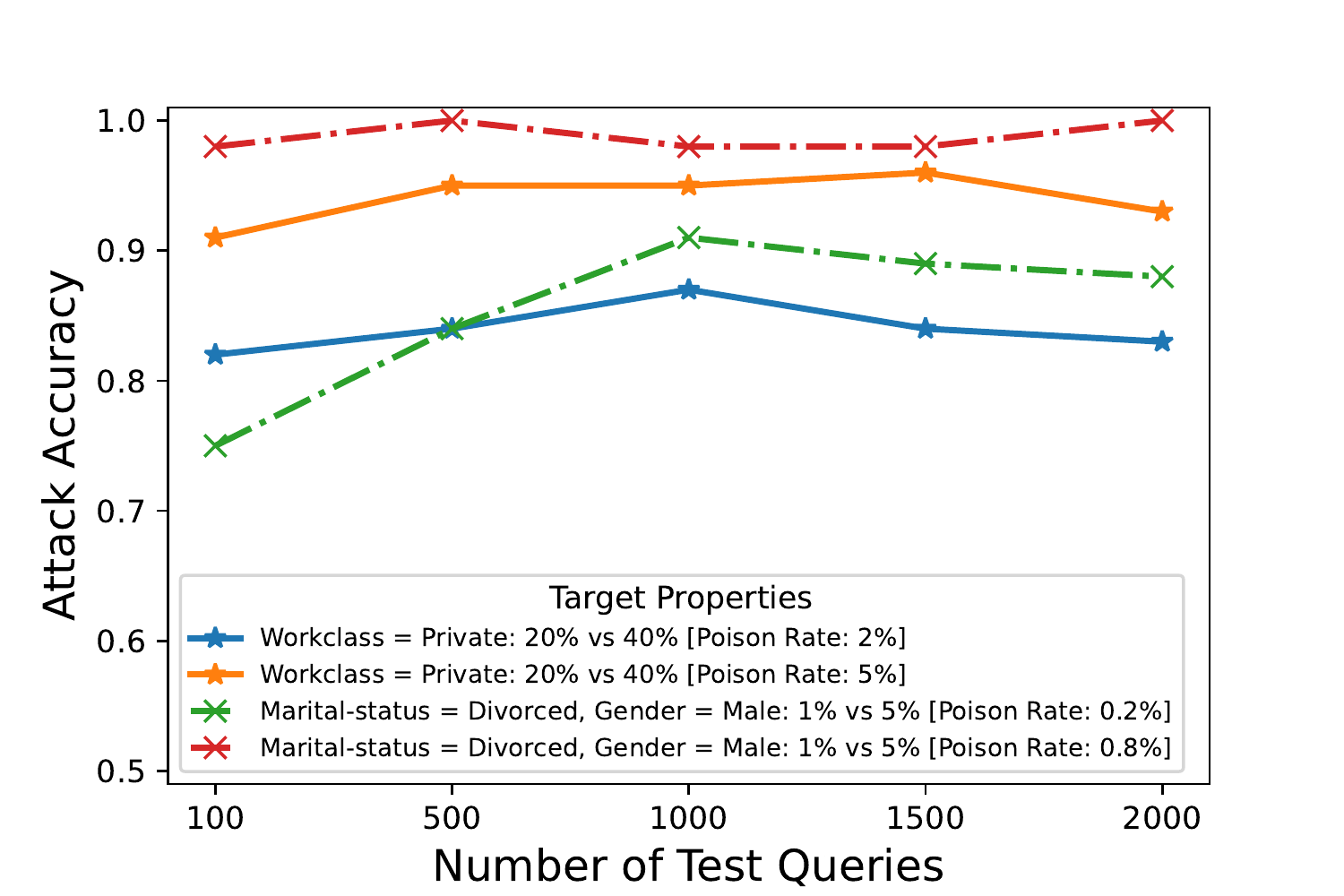}%
    	\caption{\centering 
    	Attack accuracy by  the number of test queries. Higher poisoning rate requires fewer queries to successfully distinguish between the fractions.}
	\label{fig:TQs}
    \end{minipage}
    


\end{figure*}

\myparagraph{Training set size} 
In this experiment we fix the model architecture to 2NN and vary the size of the training dataset in Figure \ref{fig:TrainSS}  to understand its impact on  the attack accuracy.
We observe that the attack accuracy improves with larger training sets. With more samples, the shadow models are able to learn the logit distributions better, and, consequently  the attack accuracy increases. We also observe that when the poisoning rate is high, the models achieve high success with  fewer training samples. Our explanation is that with more poisoning the variance of the logit distribution reduces and, as a result, we can fit a Gaussian distribution on the logit values even with just a few samples of the target property.

\myparagraph{Number of shadow models}
Next, we examine the effect on attack accuracy by varying the number of shadow models to infer the logit distribution.
We conduct our ablation on  ``Race = Black'' and ``Marital-Status = Divorced; Gender = Male'' properties from  Table \ref{tab:TP}. 
We observe that the attack accuracy for  both properties is already higher than $90\%$ with just one shadow model per fraction. At low poisoning rates, the attack accuracy improves with increase in shadow models  as  the attacker learns a better logit distribution, and consequently computes a better threshold $\threshold$ for the distinguishing test. However, as the poisoning rate is increased, for instance for target property ``Race = Black'' with poisoning rate $p = 5\%$,  we achieve a high attack accuracy of $96\%$ with only $2$ shadow models. This happens because the variance of the logit distribution shrinks with increase in poisoning rate as observed in Figure~\ref{fig:TandEResults}, and as a result fewer shadow models are enough to obtain a good representation of the distribution.

\myparagraph{\revision{Number of Test queries}} \revision{We now analyze the effect on attack accuracy by varying the set size used for querying the target model in Figure~\ref{fig:TQs}. We observe that for low poisoning rates, the attack accuracy improves and then stabilizes as the number of queries increases. The same phenomenon occurs much earlier when the poisoning rate is higher. These observations adhere to our  Claim \ref{prop:TestQueries}, where our analysis suggests fewer queries are needed at larger poisoning rates. Moreover, our analysis provides a conservative lower bound on the query set size, while in practice we  achieve similar attack accuracy with fewer queries. For instance, for target property  ``Workclass=Private'' on Adult, our analysis suggests around $260$ test queries at poisoning rate $p=5\%$, while Figure~\ref{fig:TQs} shows that 100 queries are enough to obtain the same attack success of $92\%$.}

\myparagraph{\revision{Evaluating more properties}} \revision{In our previous experiments, we selected representative properties using sensitive features in our datasets such as Race and Gender,  in line with those used in prior work~\cite{Chase21}. The target properties were chosen to cover a wide range of attributes and fractions across the four datasets. Here, we provide further evidence  that our attack strategy generalizes to a wider set of properties.  
We perform an experiment on the Adult dataset where we attack large properties by trying all possible combinations of the  features used for defining properties in Table~\ref{tab:TP}.
We observe that from all the combinations, 15  properties fall in the large category (having more than 10\% representation in the training set). We set the distinguishing test as $10\%$ vs $25\%$ with poisoning rate $p = 5\%$ and observe that \system\ still achieves an attack accuracy greater than $93\%$ across all  15 properties. This demonstrates  our attack's generalization to a wide range of properties.}

\subsection{Property Existence Evaluation}

The goal of the attacker performing  property existence  is to identify if a target property is present at all in the training  dataset or not. We target three such properties as described in Table~\ref{tab:TP}, to check for their existence in the dataset. Figure~\ref{fig:PropExst} shows the attack accuracy as a function of the amount of poisoning. Note that the  $x$ axis in Figure~\ref{fig:PropExst} represents the number of samples used for poisoning instead of the poisoning rate as used in our previous  experiments (Figures \ref{fig:PRLP},  \ref{fig:PRSP}, and \ref{fig:PRLPOPT}). The reason for this choice is that the property existence attack works with a few poisoning samples since  even a small amount of poisoning will induce a separation in the logit distributions. We reach 95\% attack accuracy  with only $8$ poisoned samples, for all three considered properties. We notice that without poisoning the property existence attack does not work very well (its success is between 29\% and 62\%), but a small amount of poisoning makes a huge difference in the attack success. A similar observation was made in Truth Serum \cite{TruthSerum}, which also showed that a small amount of poisoning improved the success of membership inference attacks.

\subsection{Comparison to previous work}
\label{sec:compare}


After evaluating the performance of our attack under multiple parameter settings, we now turn our attention to  comparison with previous work. The attack by Mahloujifar et al.~\cite{Chase21} is the only property inference attack which uses poisoning and is the closest related to ours. Mahloujifar et al. compare their attack to previous property inference attacks without poisoning~\cite{Ganju18} and show the benefits achieved by poisoning. Therefore, we compare \system\ only with~\cite{Chase21}.


The attack from \cite{Chase21} requires black-box access to the trained model, similar to \system. 
Mahloujifar et al.'s attack can be summarized in three main steps: 
\begin{itemize}
    \item Data Poisoning: The poisoned dataset $D_p$ is constructed by collecting samples with the target property and assigning them a specific label $\tlabel$. 
    
    \item Query Selection: An ensemble of $r$ models is created, where each model is trained on a random sample of 500 records with the property and 500 records  without it. The ensemble is  used to select a set of query samples $D_q$ for the distinguishing test.
    
    \item Distinguishing Test: A set of $k$ shadow models are trained per world and then queried on $D_q$. The output labels of the shadow models are  used to construct a dataset for training the attack meta-classifier model, which is then used to predict World 0 or World 1.
\end{itemize}

\noindent The  implementation from \cite{Chase21} is currently not  publicly available and we implemented the attack   with the help of the authors. We initialize their attack with target label $\tlabel = 1$ (for Data Poisoning), number of models in the ensemble $r = 100$ (for Query Selection), and the number of shadow models $k = 500$ (for Distinguishing Test), where each shadow model is trained on a random subset of $1500$ samples. The size of $D_q$ is set to $1000$ queries and the architecture used for the shadow models is a logistic regression model. These parameter choices are confirmed by the authors to be similar to those in~\cite{Chase21}. For a fair comparison we run our attack with the same  parameters, except that we use $4$ shadow models. 

\begin{figure}[h!]
 \vspace{-0.5cm}
 \centering
	\includegraphics[width=0.6\textwidth]{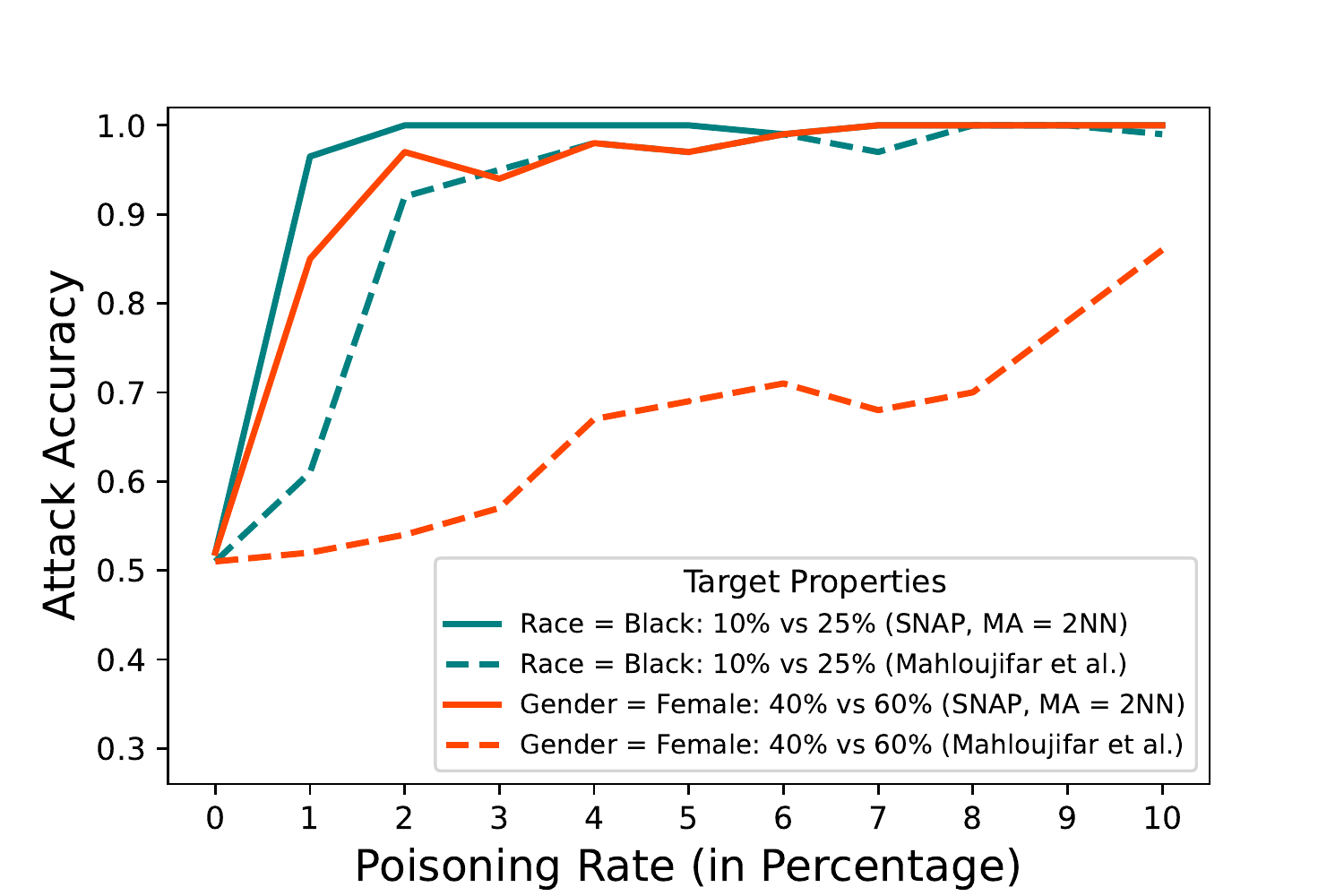}%
	\caption{Comparison of \system\ to \cite{Chase21}  with two-layered neural network as the shadow model architecture. Our attack consistently outperforms \cite{Chase21} across various poisoning rates.}
	\label{fig:Compv1}
\end{figure}

\revision{We first compare  the model confidence version of \system\ with~\cite{Chase21} for two target properties on the Census dataset: ``Gender = Female'' and ``Race = Black''. We  observe that  our attack consistently outperforms~\cite{Chase21} and requires  lower poisoning rates. For instance, for target property ``Gender = Female'', \system\ obtains $91\%$ accuracy at $3\%$ poisoning, while \cite{Chase21} obtains 57\%  accuracy at the same poisoning rate. Similarly, for target property ``Race = Black'', \system\ achieves $93\%$ accuracy at $2\%$ poisoning, while \cite{Chase21} achieves only $58\%$ accuracy. In addition, our attack takes a total of 13.6 seconds, while the Mahloujifar et al. attack needs 768.2 seconds. These results are averaged over 5 independent trials and timings are measured on a local machine with an M1 chip and 8-core CPU, as results on a GPU-enabled  machine resulted in longer run times for both attack strategies (due to the small size of the logistic regression model). We observe that \system\ is significantly faster than~\cite{Chase21} with a run time improvement  of  $56.5 \times$. 
We observe  higher attack success for \system\ when we change the model architecture to a two layer neural network (2NN) as shown in Figure~\ref{fig:Compv1}.  Our attack accuracy for both properties reaches above $96\%$ at only $2\%$ poisoning rate, while \cite{Chase21} has attack accuracy below $60\%$ for the ``Gender = Female'' property.}


\revision{We also compare our  strategy to~\cite{Chase21} on the CelebA dataset, using a distinguishing test from \cite{Chase21} to determine whether the percentage of Males in the dataset is $30\%$ or $70\%$ on the smile detection classification task. Given the poisoned dataset, our attack accuracy is higher than \cite{Chase21} with 250x fewer shadow models. At a poisoning rate of 2\%, SNAP achieves 92\% attack success on the distinguishing test, which is obtained by \cite{Chase21} at 5\% poisoning. Moreover, at 5\% poisoning, our attack achieves 100\% success, an improvement of 8\% over \cite{Chase21}. We defer the details of our setup for CelebA dataset to Appendix \ref{apndx:CompMSR}.}

\revision{As Mahloujifar et al.~\cite{Chase21} only used the class labels from the target model,  we compare it against the \system\ label-only extension described in Section~\ref{sec:ext}. 
Table~\ref{tab:LabelOnlyTab} provides the details of our comparison over multiple target properties, where we use the optimized version of our attack for large properties and choose the poisoning rate $p^*$ based on our theoretical analysis summarized in Section~\ref{sec:ext}. We observe that the label-only version of \system\ also  consistently outperforms \cite{Chase21} across all properties. For instance, for target property ``Workclass = Private'', \system\ achieves an accuracy of $94\%$ at only $1.1\%$ poisoning, while \cite{Chase21} obtains an accuracy of $56\%$ for the same poisoning rate.}  

{\begin{table}[h!]
		\centering 
		\begin{adjustbox}{max width=0.7 \textwidth}{  
				\begin{tabular}{r c c c c} 
					
					 \multicolumn{1}{c}{\multirow{1}{*}{Target Property}} & \multirow{1}{*}{Distinguishing Test} & \multirow{1}{*}{$p^*$} & \multicolumn{1}{c}{SNAP} & \multirow{1}{*}{Mahloujifar et al.\cite{Chase21}} \\ 
					
					\midrule

                     Race= Black &  $10\%$ vs $25\%$ & $3.7\%$ & $100\%$ & $97\%$\\
					
				   Gender= Female &  $30\%$ vs $50\%$ & $4.5\%$ & $98\%$  & $70\%$\\
					 
				   Workclass= Private &  $20\%$ vs $40\%$ & $1.1\%$ & $94\%$  & $56\%$\\ 

                    {Gender= Male Race= White} &  $15\%$ vs $30\%$ & $5.7\%$ & $95\%$ & $65\%$\\

				\end{tabular}
			}
		\end{adjustbox}
		\caption{ \revision{Comparison of  label-only version of \system\ with \cite{Chase21} based on our optimal poisoning rate $p^*$. \system\ consistently outperforms \cite{Chase21} across various target properties. 
		}} 
		\label{tab:LabelOnlyTab}
	\end{table}
}

\subsection{Estimating size of target property}
We conducted our property estimation for eight subpopulations whose actual fractions ($t^{*}$) range from medium to large. In these experiments, the target model and shadow model architectures are two-layer neural networks. Our estimations for each property at a given poisoning rate are averaged over 5 trials where we make 2000 queries to the target model with 2 shadow models per estimated $\hat{t}$. For these experiments, we search for $t^{*}$ over the discrete interval $[0, 1]$, incremented by $0.001$ and initialize $\hat{t}$ as $0.5$. Additionally, we stop when either 95\% of the shadow models' logits distribution falls within the bounds of the target model's logit distribution \textit{or} when the algorithm has made 6 iterations. We present the results for medium and large subpopulation estimations in Tables 4 and 5. Each iteration of the estimation algorithm took 74 seconds and 16 seconds on an M1 CPU for Census and Adult, respectively. With our experimental setup, the maximum possible running time for a single estimation is 7 minutes and 24 seconds on Census and 1 minute and 36 seconds on Adult. 

In a similar fashion to the distinguishing test, the property estimation algorithm achieves better success as poisoning increases. For medium properties, our method requires up to 1\% poisoning to achieve estimates within 1\% - 10\% of the true fraction, $t^{*}$. For larger properties, such as ``Gender = Female'' and ``Workclass = Private'', we require up to 5\% poisoning to achieve estimates within 0\% - 5\% of $t^{*}$.

\begin{table}[h!]
\centering
\resizebox{0.7\columnwidth}{!}{%
\begin{tabular}{rcrrr}

\multicolumn{1}{c}{\multirow{3}{*}{Target Property}} & \multirow{3}{*}{$t^*$} & \multicolumn{3}{c}{Our Estimation} \\ \cmidrule{3-5}
             &  & \multicolumn{1}{c}{0\%} & \multicolumn{1}{c}{0.5\%} & \multicolumn{1}{c}{1\%} \\ \midrule
Industry = Construction & 3.0\%    & 32.6\%   & 3.7\%  & {\bf 3.1\%}   \\
Education = Bachelors                   & 10.0\%  & 58.3\%      & {\bf 9.9\%}     & 10.4\%     \\ 

Gender = Female, Occupation = Sales      & 3.9\%   & 24.9\%      & 9.9\%     & {\bf 4.3\%}     \\

Gender = Male; Marital-Status = Divorced & 5.4\%   & 33.7\%     & {\bf 5.8\%}    & 6.1\%     \\ 
\end{tabular}%
}
\caption{Estimated $t$ values from our property estimation algorithm on medium target properties at varying poisoning rates.}
\label{tab:estimation-medium}
\end{table}

\begin{table}[h!]
\centering
\resizebox{0.7\columnwidth}{!}{%
\begin{tabular}{rcrrrr}

\multicolumn{1}{c}{\multirow{3}{*}{Target Property}} & \multirow{3}{*}{$t^*$} & \multicolumn{4}{c}{Our Estimation} \\ \cmidrule{3-6}
             &  & \multicolumn{1}{c}{0\%} & \multicolumn{1}{c}{1\%} & \multicolumn{1}{c}{3\%} & \multicolumn{1}{c}{5\%} \\ \midrule
Gender = Female             & 52.1\% & 30.3\%  & 41.9\% & 45.0\% & {\bf 50.0\%}   \\ 
Race = Black                & 10.2\%  & 17.5\%  & {\bf 10.3\%} & 9.2\%  & 9.3\%  \\ 

Workclass = Private          & 40.0\%   & 50.0\%    & 50.0\%   & 50.0\%   & {\bf 40.0\%}   \\
Race = White, Gender = Male  & 43.0\%  & 31.5\%  & 25.9\% & 50.0\% & {\bf 45.0\%} \\ 
\end{tabular}%
}
\caption{Performance of our property estimation algorithm on large target properties by poisoning rate.}
\label{tab:estimation-medium-large}
\end{table}

\revision{We also empirically measure the robustness of our estimation algorithm to overpoisoning. In some settings, the adversary may not know what poisoning rate they should use prior to running the estimation. Because of this, the adversary may choose a larger poisoning rate than the optimal one. Poisoning with rates  higher than those shown in Tables \ref{tab:estimation-medium} and \ref{tab:estimation-medium-large} still results in effective property size estimation with low estimation error. For instance, the properties ``Marital-Status = Divorced; Gender = Male'' (on Adult) and ``Race = Black'' (on Census) make up 5.4\% and 10.2\% of their datasets, respectively. Although a poisoning rate of 0.5\% is optimal for ``Marital-Status = Divorced; Gender = Male,'' our average estimate stays within 1.5\% of $t^{*}$ when we set the poisoning rate to 5.0\% (ten times higher than optimal). Similarly, the optimal poisoning rate for ``Race = Black'' is 1\%, but our estimate remains within 0.9\% of $t^{*}$ when increasing the poisoning rate by ten times to 10\%. The attack is thus not very sensitive to the exact selection of the poisoning rate.}

\revision{Finally, we analyze how changing the stopping condition of our estimation algorithm impacts the accuracy of the estimation. In the default setting, we stop when  95\% of the shadow models' logits distribution overlap with the target model's logit distribution. We vary the 95\% threshold, using   70\%, 90\%, and 99\%, for property ``Marital-Status = Divorced; Gender = Male'' ($t^{*}=5.4\%$) at 5\%  poisoning rate. Our average estimates over 5 trials are 13.7\%, 7.5\%, and 6.2\%, respectively, compared to the 5.8\% estimate for a threshold of 95\%.  Although the estimation accuracy is lower, using a smaller fraction for the stopping condition allows the estimation algorithm to converge in fewer iterations.}

\section{Discussion and Conclusion}

We introduce an efficient property inference attack motivated by sound theoretical analysis of the effects of poisoning on model confidence distributions. Our  attack, \system, outperforms prior work \cite{Ganju18, Chase21, DistributionInference} in multiple settings while requiring several orders of magnitude fewer shadow models. This resulted in a $56.5 \times$ speed increase when compared to \cite{Chase21} on the Census dataset and a higher attack success at lower poisoning rates than \cite{Chase21}. \system\ achieves above 90\% attack success  with only 8 poisoned samples on small properties, 0.6\% poisoning on medium properties, and 1.5\% poisoning on large properties.


 We also introduce several extensions to our \system\ framework.
Our property existence attack extends \system\ to be a generalization of membership inference, where we are able to determine if a group of individuals with a certain property have been used in training the target model. The label-only extension only requires the target model's predicted labels and outperforms attacks from previous work~\cite{Chase21}. The property size estimation attack generalizes \system\ by requiring no prior knowledge of possible fractions $t_{0}$ and $t_{1}$. It  makes precise estimates of property proportion in the training set with low poisoning rates, while requiring exponentially fewer shadow models than previous work~\cite{DistributionInference} and \cite{Chase21}.

Next, we discuss several aspects of our attacks and directions for future work.

\myparagraph{\revision{Attack Configuration}} \revision{The parameters for our attack can be selected by following the theoretical analysis of model confidences under poisoning, described in Theorem \ref{thm:GaussianDist}. Our bounds, depicted in Figure \ref{fig:TandEResults}, provide good estimates for how the poisoning rate affects the mean and variance of the target model's logit distribution. Consequently, as described in Section \ref{sec:analysis}, choosing a poisoning rate $p$ for which the theoretical variance of both worlds is below a fixed threshold (e.g., 0.15) is a valid strategy. Another strategy is to choose the poisoning rate $p$ such the difference between the theoretical logit means of the two worlds is above a fixed threshold. For the label-only extension described in Section~\ref{sec:ext}, our principled approach of choosing an optimal poisoning rate gives a good strategy for consistently obtaining high attack accuracy.}


\myparagraph{\revision{Selecting Target Properties}} \revision{Given that our attack requires poisoning, the adversary needs to commit to a target property and poisoning points before the target ML model is trained. We observed that our attack not only builds an accurate distinguishing test for the target property itself, but can also be used to infer information about underlying sub-properties. We have exploited this observation to design the optimized version of our attack described in Section~\ref{sec:expPropInf}.
Thus, an attacker has the potential to  infer information on multiple sub-properties by poisoning a larger target property during training.  We leave formulating a poisoning strategy that maximizes the attack accuracy on sub-properties of the target property as future work.}


\myparagraph{Defenses against Property Inference} Differential privacy was explicitly designed to make it possible to reveal statistical properties of a dataset~\cite{OriginalDP}, and thus is not intended to provide defense against property inference. Prior work~\cite{Chase21} confirmed empirically that differentially private training algorithms do not defend against property inference attacks.

\revision{To empirically test this, we trained our target models using DP-Adam and ran our attack on two properties from the Census and Adult datasets (``Race = White, Gender = Male'' from Adult and ``Race = Black'' from Census). Overall, \system\ is able to achieve high success even when the target model has been trained with differential privacy as shown in Table~\ref{tab:DPAdam} (Appendix ~\ref{apndx:DP}). The configurations of these experiments are also detailed in Appendix~\ref{apndx:DP}. }






The attack also performs better as the number of queries increases, and therefore bounding the number of queries by user is a potential strategy to mitigate these attacks. We also show that our attack success rate improves as the amount of poisoning increases, suggesting that applying poisoning defenses \cite{Ma19, CertifiedDefenses, ImprovedCertifiedDefenses} may help prevent our attacks. We leave a thorough evaluation of defenses for property inference or a proof of defense impossibility  for future work. 

\myparagraph{Property Inference as an Auditing Tool} In addition to revealing sensitive information about training datasets of ML models, property inference has the potential to be used as a tool for auditing the fairness of  ML models. If a company shares their model with a third party, they would be able to determine the demographics of the dataset used to train the model using different \system\ attack variants. This way, an auditor could efficiently determine whether the dataset used to train this company's model contains fair representations of its constituent properties. We believe that adapting property inference for auditing of ML fairness is a promising direction for future work. 

\section*{Acknowledgements}
\noindent
We thank Saeed Mahloujifar for providing guidelines for implementing the attack from~\cite{Chase21}. We thank Nicholas Carlini for helpful feedback on a draft of this paper. Alina Oprea was supported partially by NSF grant CNS-2120603. Jonathan Ullman was supported by NSF grants CCF-1750640, CNS-1816028, and CNS-2120603.



\bibliographystyle{IEEEtran}
\bibliography{refs/main_short,refs/alina,refs/refs_advml}

\begin{thebibliography}{10}
\providecommand{\url}[1]{#1}
\csname url@samestyle\endcsname
\providecommand{\newblock}{\relax}
\providecommand{\bibinfo}[2]{#2}
\providecommand{\BIBentrySTDinterwordspacing}{\spaceskip=0pt\relax}
\providecommand{\BIBentryALTinterwordstretchfactor}{4}
\providecommand{\BIBentryALTinterwordspacing}{\spaceskip=\fontdimen2\font plus
\BIBentryALTinterwordstretchfactor\fontdimen3\font minus
  \fontdimen4\font\relax}
\providecommand{\BIBforeignlanguage}[2]{{%
\expandafter\ifx\csname l@#1\endcsname\relax
\typeout{** WARNING: IEEEtran.bst: No hyphenation pattern has been}%
\typeout{** loaded for the language `#1'. Using the pattern for}%
\typeout{** the default language instead.}%
\else
\language=\csname l@#1\endcsname
\fi
#2}}
\providecommand{\BIBdecl}{\relax}
\BIBdecl

\bibitem{Ganju18}
\BIBentryALTinterwordspacing
K.~Ganju, Q.~Wang, W.~Yang, C.~A. Gunter, and N.~Borisov, ``Property inference
  attacks on fully connected neural networks using permutation invariant
  representations,'' in \emph{Proceedings of the 2018 ACM SIGSAC Conference on
  Computer and Communications Security}, ser. CCS '18.\hskip 1em plus 0.5em
  minus 0.4em\relax New York, NY, USA: Association for Computing Machinery,
  2018, p. 619–633. [Online]. Available:
  \url{https://doi.org/10.1145/3243734.3243834}
\BIBentrySTDinterwordspacing

\bibitem{Zhang21}
\BIBentryALTinterwordspacing
W.~Zhang, S.~Tople, and O.~Ohrimenko, ``Leakage of dataset properties in
  {Multi-Party} machine learning,'' in \emph{30th USENIX Security Symposium
  (USENIX Security 21)}.\hskip 1em plus 0.5em minus 0.4em\relax USENIX
  Association, Aug. 2021, pp. 2687--2704. [Online]. Available:
  \url{https://www.usenix.org/conference/usenixsecurity21/presentation/zhang-wanrong}
\BIBentrySTDinterwordspacing

\bibitem{Chase21}
S.~Mahloujifar, E.~Ghosh, and M.~Chase, ``Property inference from poisoning,''
  in \emph{2022 IEEE Symposium on Security and Privacy (SP)}, 2022, pp.
  1120--1137.

\bibitem{Ateniese15}
\BIBentryALTinterwordspacing
G.~Ateniese, L.~V. Mancini, A.~Spognardi, A.~Villani, D.~Vitali, and G.~Felici,
  ``Hacking smart machines with smarter ones: How to extract meaningful data
  from machine learning classifiers,'' \emph{Int. J. Secur. Netw.}, vol.~10,
  no.~3, p. 137–150, sep 2015. [Online]. Available:
  \url{https://doi.org/10.1504/IJSN.2015.071829}
\BIBentrySTDinterwordspacing

\bibitem{DistributionInference}
A.~Suri and D.~Evans, ``Formalizing and estimating distribution inference
  risks,'' \emph{Proceedings on Privacy Enhancing Technologies}, 2022.

\bibitem{shokri2017membership}
R.~Shokri, M.~Stronati, C.~Song, and V.~Shmatikov, ``Membership inference
  attacks against machine learning models,'' in \emph{2017 IEEE Symposium on
  Security and Privacy (SP)}.\hskip 1em plus 0.5em minus 0.4em\relax IEEE,
  2017, pp. 3--18.

\bibitem{Dua:2019}
\BIBentryALTinterwordspacing
D.~Dua and C.~Graff, ``{UCI} machine learning repository,'' 2017. [Online].
  Available: \url{http://archive.ics.uci.edu/ml}
\BIBentrySTDinterwordspacing

\bibitem{CelebA}
Z.~Liu, P.~Luo, X.~Wang, and X.~Tang, ``Deep learning face attributes in the
  wild,'' in \emph{Proceedings of International Conference on Computer Vision
  (ICCV)}, December 2015.

\bibitem{DinurN03}
I.~Dinur and K.~Nissim, ``Revealing information while preserving privacy,'' in
  \emph{Proceedings of the 22nd ACM Symposium on Principles of Database
  Systems}, ser. PODS '03.\hskip 1em plus 0.5em minus 0.4em\relax ACM, 2003,
  pp. 202--210.

\bibitem{SecretSharer}
N.~Carlini, C.~Liu, {\'U}.~Erlingsson, J.~Kos, and D.~Song, ``The secret
  sharer: Evaluating and testing unintended memorization in neural networks,''
  in \emph{28th USENIX Security Symposium (USENIX Security 19)}.\hskip 1em plus
  0.5em minus 0.4em\relax USENIX Association, Aug. 2019.

\bibitem{MemorizationGPT2}
N.~Carlini, F.~Tramer, E.~Wallace, M.~Jagielski, A.~Herbert-Voss, K.~Lee,
  A.~Roberts, T.~Brown, D.~Song, U.~Erlingsson, A.~Oprea, and C.~Raffel,
  ``Extracting training data from large language models,'' in \emph{30th
  {USENIX} Security Symposium ({USENIX} Security 2021)}, 2021.

\bibitem{InformedAdv}
\BIBentryALTinterwordspacing
B.~Balle, G.~Cherubin, and J.~Hayes, ``Reconstructing training data with
  informed adversaries,'' \emph{CoRR}, vol. abs/2201.04845, 2022. [Online].
  Available: \url{https://arxiv.org/abs/2201.04845}
\BIBentrySTDinterwordspacing

\bibitem{FL_Disaggregation}
\BIBentryALTinterwordspacing
M.~Lam, G.-Y. Wei, D.~Brooks, V.~J. Reddi, and M.~Mitzenmacher, ``Gradient
  disaggregation: Breaking privacy in federated learning by reconstructing the
  user participant matrix,'' in \emph{Proceedings of the 38th International
  Conference on Machine Learning}, ser. Proceedings of Machine Learning
  Research, M.~Meila and T.~Zhang, Eds., vol. 139.\hskip 1em plus 0.5em minus
  0.4em\relax PMLR, 18--24 Jul 2021, pp. 5959--5968. [Online]. Available:
  \url{https://proceedings.mlr.press/v139/lam21b.html}
\BIBentrySTDinterwordspacing

\bibitem{FL_Privacy}
F.~Boenisch, A.~Dziedzic, R.~Schuster, A.~S. Shamsabadi, I.~Shumailov, and
  N.~Papernot, ``When the curious abandon honesty: Federated learning is not
  private,'' 2021.

\bibitem{Homer+08}
N.~Homer, S.~Szelinger, M.~Redman, D.~Duggan, W.~Tembe, J.~Muehling, J.~V.
  Pearson, D.~A. Stephan, S.~F. Nelson, and D.~W. Craig, ``Resolving
  individuals contributing trace amounts of {DNA} to highly complex mixtures
  using high-density {SNP} genotyping microarrays,'' \emph{PLoS genetics},
  vol.~4, no.~8, p. e1000167, 2008.

\bibitem{yeom2018privacy}
S.~Yeom, I.~Giacomelli, M.~Fredrikson, and S.~Jha, ``Privacy risk in machine
  learning: Analyzing the connection to overfitting,'' in \emph{2018 IEEE 31st
  Computer Security Foundations Symposium (CSF)}.\hskip 1em plus 0.5em minus
  0.4em\relax IEEE, 2018, pp. 268--282.

\bibitem{Sablayrolles19}
A.~Sablayrolles, M.~Douze, C.~Schmid, Y.~Ollivier, and H.~Jégou., ``White-box
  vs black-box: Bayes optimal strategies for membership inference.''\hskip 1em
  plus 0.5em minus 0.4em\relax In International Conference on Machine
  Learning., 2019.

\bibitem{Long20}
Y.~Long, L.~Wang, D.~Bu, V.~Bindschaedler, X.~Wang, H.~Tang, C.~A. Gunter, and
  K.~Chen., ``A pragmatic approach to membership inferences on machine learning
  models.''\hskip 1em plus 0.5em minus 0.4em\relax In IEEE European Symposium
  on Security and Privacy (EuroS\&P), 2020.

\bibitem{Song21}
L.~Song and P.~Mittal., ``Systematic evaluation of privacy risks of machine
  learning models.''\hskip 1em plus 0.5em minus 0.4em\relax In 30th USENIX
  Security Symposium, 2021.

\bibitem{Jayaraman21}
B.~Jayaraman, L.~Wang, D.~Evans, and Q.~Gu., ``Revisiting membership inference
  under realistic assumptions.''\hskip 1em plus 0.5em minus 0.4em\relax In
  Proceedings on Privacy Enhancing Technologies (PoPETs), 2021.

\bibitem{Choo21}
C.~A. Choquette-Choo, F.~Tramer, N.~Carlini, and N.~Papernot, ``Label-only
  membership inference attacks,'' in \emph{Proceedings of the 38th
  International Conference on Machine Learning}.\hskip 1em plus 0.5em minus
  0.4em\relax PMLR, 2021.

\bibitem{EnhancedMembership}
J.~Ye, A.~Maddi, S.~K. Murakonda, and R.~Shokri, ``Enhanced membership
  inference attacks against machine learning models.''\hskip 1em plus 0.5em
  minus 0.4em\relax arXiv, 2021.

\bibitem{LiRA}
N.~Carlini, S.~Chien, M.~Nasr, S.~Song, A.~Terzis, and F.~Tramer, ``Membership
  inference attacks from first principles,'' in \emph{2022 2022 IEEE Symposium
  on Security and Privacy (SP) (SP)}.\hskip 1em plus 0.5em minus 0.4em\relax
  Los Alamitos, CA, USA: IEEE Computer Society, may 2022, pp. 1519--1519.

\bibitem{Biggio2012PoisoningAA}
B.~Biggio, B.~Nelson, and P.~Laskov, ``Poisoning attacks against support vector
  machines,'' in \emph{Proceedings of the 29th International Conference on
  International Conference on Machine Learning, ICML}, 2012.

\bibitem{xiao2015feature}
H.~Xiao, B.~Biggio, G.~Brown, G.~Fumera, C.~Eckert, and F.~Roli, ``Is feature
  selection secure against training data poisoning?'' in \emph{International
  Conference on Machine Learning}, 2015, pp. 1689--1698.

\bibitem{ManipulatingML}
M.~Jagielski, A.~Oprea, B.~Biggio, C.~Liu, C.~Nita-Rotaru, and B.~Li,
  ``Manipulating machine learning: Poisoning attacks and countermeasures for
  regression learning,'' in \emph{2018 IEEE Symposium on Security and Privacy
  (SP)}, 2018, pp. 19--35.

\bibitem{NKS06}
J.~Newsome, B.~Karp, and D.~Song, ``Paragraph: Thwarting signature learning by
  training maliciously,'' in \emph{Recent Advances in Intrusion Detection},
  2006.

\bibitem{koh2017understanding}
P.~W. Koh and P.~Liang, ``Understanding black-box predictions via influence
  functions,'' in \emph{Proceedings of the 34th International Conference on
  Machine Learning-Volume 70}.\hskip 1em plus 0.5em minus 0.4em\relax JMLR.
  org, 2017, pp. 1885--1894.

\bibitem{suciu2018does}
O.~Suciu, R.~Marginean, Y.~Kaya, H.~Daume~III, and T.~Dumitras, ``When does
  machine learning $\{$FAIL$\}$? generalized transferability for evasion and
  poisoning attacks,'' in \emph{27th {USENIX} Security Symposium}, 2018, pp.
  1299--1316.

\bibitem{geiping2020witches}
\BIBentryALTinterwordspacing
J.~Geiping, L.~H. Fowl, W.~R. Huang, W.~Czaja, G.~Taylor, M.~Moeller, and
  T.~Goldstein, ``Witches' brew: Industrial scale data poisoning via gradient
  matching,'' in \emph{International Conference on Learning Representations},
  2021. [Online]. Available: \url{https://openreview.net/forum?id=01olnfLIbD}
\BIBentrySTDinterwordspacing

\bibitem{GLDG19}
T.~{Gu}, K.~{Liu}, B.~{Dolan-Gavitt}, and S.~{Garg}, ``Badnets: Evaluating
  backdooring attacks on deep neural networks,'' \emph{IEEE Access}, 2019.

\bibitem{CLLLS17}
X.~Chen, C.~Liu, B.~Li, K.~Lu, and D.~Song, ``Targeted backdoor attacks on deep
  learning systems using data poisoning,'' 2017.

\bibitem{SubpopulationPoisoning}
M.~Jagielski, G.~Severi, N.~P. Harger, and A.~Oprea, ``Subpopulation data
  poisoning attacks,'' in \emph{Proceedings of the ACM Conference on Computer
  and Communications Security}, ser. CCS, 2021.

\bibitem{Ma19}
J.~H. Yuzhe~Ma, Xiaojin~Zhu, ``Data poisoning against differentially-private
  learners: Attacks and defenses.''\hskip 1em plus 0.5em minus 0.4em\relax In
  Proceedings of the 28th International Joint Conference on Artificial
  Intelligence (IJCAI), 2019.

\bibitem{AuditingDP}
M.~Jagielski, J.~Ullman, and A.~Oprea, ``Auditing differentially private
  machine learning: How private is private {SGD?}'' in \emph{Advances in Neural
  Information Processing Systems}, vol.~33, 2020, pp. 22\,205--22\,216.

\bibitem{nasr2021adversary}
M.~Nasr, S.~Songi, A.~Thakurta, N.~Papemoti, and N.~Carlin, ``Adversary
  instantiation: Lower bounds for differentially private machine learning,'' in
  \emph{2021 IEEE Symposium on Security and Privacy (SP)}.\hskip 1em plus 0.5em
  minus 0.4em\relax IEEE, 2021, pp. 866--882.

\bibitem{TruthSerum}
F.~Tramèr, R.~Shokri, A.~S. Joaquin, H.~Le, M.~Jagielski, S.~Hong, and
  N.~Carlini, ``{Truth Serum}: Poisoning machine learning models to reveal
  their secrets,'' in \emph{ACM Computer and Communications Security (CCS)},
  2022.

\bibitem{FeatureLeakage}
\BIBentryALTinterwordspacing
L.~Melis, C.~Song, E.~D. Cristofaro, and V.~Shmatikov, ``Exploiting unintended
  feature leakage in collaborative learning,'' in \emph{2019 {IEEE} Symposium
  on Security and Privacy, {SP} 2019, San Francisco, CA, USA, May 19-23,
  2019}.\hskip 1em plus 0.5em minus 0.4em\relax {IEEE}, 2019, pp. 691--706.
  [Online]. Available: \url{https://doi.org/10.1109/SP.2019.00029}
\BIBentrySTDinterwordspacing

\bibitem{PI_GAN}
J.~Zhou, Y.~Chen, C.~Shen, and Y.~Zhang, ``Property inference attacks against
  {GANs},'' in \emph{Proceedings of Network and Distributed System Security},
  ser. NDSS, 2022.

\bibitem{Inference_GNN}
Z.~Zhang, M.~Chen, M.~Backes, Y.~Shen, and Y.~Zhang, ``Inference attacks
  against graph neural networks,'' in \emph{31st USENIX Security Symposium
  (USENIX Security 22)}, 2022.

\bibitem{pmlr-v20-biggio11}
\BIBentryALTinterwordspacing
B.~Biggio, B.~Nelson, and P.~Laskov, ``Support vector machines under
  adversarial label noise,'' in \emph{Proceedings of the Asian Conference on
  Machine Learning}, ser. Proceedings of Machine Learning Research, C.-N. Hsu
  and W.~S. Lee, Eds., vol.~20.\hskip 1em plus 0.5em minus 0.4em\relax South
  Garden Hotels and Resorts, Taoyuan, Taiwain: PMLR, 14--15 Nov 2011, pp.
  97--112. [Online]. Available:
  \url{https://proceedings.mlr.press/v20/biggio11.html}
\BIBentrySTDinterwordspacing

\bibitem{OriginalDP}
C.~Dwork, F.~McSherry, K.~Nissim, and A.~Smith, ``Calibrating noise to
  sensitivity in private data analysis,'' in \emph{Theory of Cryptography},
  S.~Halevi and T.~Rabin, Eds.\hskip 1em plus 0.5em minus 0.4em\relax Berlin,
  Heidelberg: Springer Berlin Heidelberg, 2006, pp. 265--284.

\bibitem{CertifiedDefenses}
\BIBentryALTinterwordspacing
J.~Steinhardt, P.~W.~W. Koh, and P.~S. Liang, ``Certified defenses for data
  poisoning attacks,'' in \emph{Advances in Neural Information Processing
  Systems}, I.~Guyon, U.~V. Luxburg, S.~Bengio, H.~Wallach, R.~Fergus,
  S.~Vishwanathan, and R.~Garnett, Eds., vol.~30.\hskip 1em plus 0.5em minus
  0.4em\relax Curran Associates, Inc., 2017. [Online]. Available:
  \url{https://proceedings.neurips.cc/paper/2017/file/9d7311ba459f9e45ed746755a32dcd11-Paper.pdf}
\BIBentrySTDinterwordspacing

\bibitem{ImprovedCertifiedDefenses}
\BIBentryALTinterwordspacing
W.~Wang, A.~J. Levine, and S.~Feizi, ``Improved certified defenses against data
  poisoning with ({Deterministic Finite Aggregation},'' in \emph{Proceedings of
  the 39th International Conference on Machine Learning}, ser. Proceedings of
  Machine Learning Research, K.~Chaudhuri, S.~Jegelka, L.~Song, C.~Szepesvari,
  G.~Niu, and S.~Sabato, Eds., vol. 162.\hskip 1em plus 0.5em minus 0.4em\relax
  PMLR, 17--23 Jul 2022, pp. 22\,769--22\,783. [Online]. Available:
  \url{https://proceedings.mlr.press/v162/wang22m.html}
\BIBentrySTDinterwordspacing

\bibitem{opacus}
A.~Yousefpour, I.~Shilov, A.~Sablayrolles, D.~Testuggine, K.~Prasad, M.~Malek,
  J.~Nguyen, S.~Ghosh, A.~Bharadwaj, J.~Zhao, G.~Cormode, and I.~Mironov,
  ``Opacus: {U}ser-friendly differential privacy library in {PyTorch},''
  \emph{arXiv preprint arXiv:2109.12298}, 2021.

\bibitem{AdamOptimizer}
\BIBentryALTinterwordspacing
D.~P. Kingma and J.~Ba, ``Adam: {A} method for stochastic optimization,'' in
  \emph{3rd International Conference on Learning Representations, {ICLR} 2015,
  San Diego, CA, USA, May 7-9, 2015, Conference Track Proceedings}, Y.~Bengio
  and Y.~LeCun, Eds., 2015. [Online]. Available:
  \url{http://arxiv.org/abs/1412.6980}
\BIBentrySTDinterwordspacing

\end{thebibliography}

\appendices
\section{Attack Analysis} \label{apdx:AttackAnalysis}

In this section, we provide proofs for the theorems and claims presented in Section \ref{sec:analysis}. 


\subsection{Effect of Poisoning on Logit Distribution} \label{proof:LogitDist}
%

%

\LogitDist*

\begin{proof}
By the logit definition for binary classification, and our assumption on the classifier, we can write: 
\begin{equation} \label{eqn:tgt_logit}
\resizebox{.42\textwidth}{!}{$\plogitval{x}{\tlabel} = \log\left[\frac{\fm(x)_{\tlabel}}{\fm(x)_{\vlabel}}\right] = \log \left[\frac{\Pr[\Tilde{\rv{Y}} = \tlabel | \Tilde{\rv{X}} = x]}{ \Pr[\Tilde{\rv{Y}} = \vlabel | \Tilde{\rv{X}} = x]}\right]$}
\end{equation}
where all probabilities are over the poisoned distribution.


Then we just need to compute the probability $\Pr[\Tilde{\rv{Y}} = \vlabel | \Tilde{\rv{X}} = x]$. The proof is similar to that of Mahloujifar et al.~\cite{Chase21}, but is adapted to our attack. We write the event $E_p$ for the event where an example is sampled from the poisoned distribution, which happens with probability $p$, and $E_c$ for the complementary event where an example is sampled from the clean distribution. Then we have: 
\begin{flalign*}
    \Pr[\Tilde{\rv{Y}} = \vlabel | \Tilde{\rv{X}} = x] =  \Pr[\Tilde{\rv{Y}} = \vlabel |\Tilde{\rv{X}} = x\wedge E_c] 
    \cdot \Pr[E_c| \Tilde{\rv{X}} = x]  &&\\\nonumber + \Pr[\Tilde{\rv{Y}} = \vlabel | \Tilde{\rv{X}} = x\wedge E_p]  \cdot \Pr[E_p |\Tilde{\rv{X}} = x]
\end{flalign*}
Note that $\Pr[\Tilde{Y} = \vlabel |\Tilde{X} = x \wedge E_p] = 0$, as  poisoned samples $x \leftarrow \dist_p$ always have the associated label $\tlabel$.  We can then re-write the above equation as: 
\begin{equation} \label{eqn: v|x}
    \resizebox{.48\textwidth}{!}{$\Pr[\Tilde{\rv{Y}} = \vlabel | \Tilde{\rv{X}} = x] = \Pr[\Tilde{\rv{Y}} = \vlabel |\Tilde{\rv{X}} = x \wedge E_c] \cdot \Pr[E_c|\Tilde{\rv{X}} = x] $}
\end{equation}
%
We compute  $\Pr[E_c|\Tilde{X} = x]$ using Bayes' theorem as:
\begin{equation}
\label{eqn:GE|X}
    \Pr[E_c|\Tilde{\rv{X}} = x] = \frac{\Pr[\Tilde{\rv{X}} = x | E_c] \cdot \Pr[E_c]}{\Pr[\Tilde{\rv{X}} =x ]}
\end{equation} 

We now compute the numerator of the above equation, relative to the probabilities in the clean distribution:
%
\begin{flalign}
\label{eq:bayes1}
\nonumber \Pr[E_c]\cdot \Pr[\Tilde{\rv{X}} = x | E_c] = (1-p) \cdot \Pr[\rv{X} = x] 
\\\nonumber = (1-p) \cdot \Pr[\rv{X} =x \wedge f(x)=1 \wedge \rv{Y}=v]
\\\nonumber =  \Pr[f(x)=1] \cdot\Pr[\rv{Y}=v|f(x)=1]
\\\nonumber \cdot \Pr[\rv{X} = x|\rv{Y}=v \wedge f(x)=1] \cdot (1-p)  
\\ = (1-p) t \pi_v \Pr[\rv{X} = x|\rv{Y}=v \wedge f(x)=1]
\end{flalign}

Similarly, we can rewrite the denominator as follows:
%
\begin{align}
\label{eq:bayes2}
\nonumber \Pr[\Tilde{\rv{X}} = x] = \Pr[\Tilde{\rv{X}} = x | E_c] \cdot \Pr[E_c] 
\\\nonumber+ \Pr[\Tilde{\rv{X}} = x | E_p] \cdot \Pr[E_p] 
\\=[t \pi_v (1-p) + p] 
\cdot \Pr[\rv{X} = x|\rv{Y}=\vlabel \wedge f(x)=1]
\end{align}
Substituting Eqn. (\ref{eq:bayes1}) and (\ref{eq:bayes2}) into Eqn. (\ref{eqn:GE|X}), we obtain:
\begin{align} \label{eq:bayes3}
    \Pr[E_c|\Tilde{\rv{X}} = x] =  \frac{t \pi_v  (1-p)}{t \pi_v  (1-p)+p}
\end{align}
Now substituting Eqn.(\ref{eq:bayes3}) back into Eqn. (\ref{eqn: v|x}), we get:
%
\begin{flalign} \label{eqn:v|xintermsofpt}
\nonumber \Pr[\Tilde{\rv{Y}} = \vlabel | \Tilde{\rv{X}} = x] = 
\frac{t \pi_v (1-p)}{p+t\pi_v (1-p)} 
\\\nonumber\cdot \Pr[\Tilde{\rv{Y}} = \vlabel |\Tilde{\rv{X}} = x \wedge E_c]
\\ =  \frac{t \pi_v (1-p)}{p+t\pi_v (1-p)} \cdot \Pr[\rv{Y} = \vlabel | \rv{X} = x]
\end{flalign}
Similarly, we can then calculate  probability  $\Pr[\Tilde{\rv{Y}} = \tlabel | \Tilde{\rv{X}} = x]$ using Eqn. (\ref{eqn:v|xintermsofpt}) as follows:
%
\begin{flalign} \label{eqn:t|xintermsofpt}
    \nonumber \Pr[\Tilde{\rv{Y}} = \tlabel | \Tilde{\rv{X}} = x] = 1 - \Pr[\Tilde{\rv{Y}} = \vlabel | \Tilde{\rv{X}} = x] 
    \\\nonumber = 1 - \frac{t \pi_v (1-p)}{p+t \pi_v (1-p)} . \Pr[\rv{Y} = \vlabel | \rv{X} = x]
    \\ = \frac{p}{p+ t \pi_v(1-p)} + \frac{t \pi_v(1-p)}{p+t \pi_v(1-p)} \Pr[\rv{Y} = \tlabel | \rv{X} = x]   
\end{flalign}
Substituting Eqn. (\ref{eqn:v|xintermsofpt}) and Eqn. (\ref{eqn:t|xintermsofpt}) back into Eqn. (\ref{eqn:tgt_logit}) and simplifying the equation, we get:
\begin{equation*}
    \plogitval{x}{\tlabel} = \log{\left[\frac{p}{t \pi_v(1-p)} + e^{\logitval{x}{\tlabel}} \left( 1+\frac{p}{t \pi_v(1-p)}\right) \right]}
\end{equation*}
\end{proof}

\GaussianDist*

\begin{proof}
We re-write Eqn.\ref{eqn:logit_eqn} as follows:
\begin{equation*}
    e^{\plogitval{x}{\tlabel}} = \frac{p}{\pi_v(1-p)t} + e^{\logitval{x}{\tlabel}} \left( 1+\frac{p}{\pi_v(1-p)t}\right)
\end{equation*}
Under the assumption that logit value $\logitval{x}{\tlabel}$ is a Gaussian random variable,  random variable $e^{\logitval{x}{\tlabel}}$ as a consequence follows log-normal distribution with mean $e^{~\mu + \sigma^2/2}$ and variance $(e^{\sigma^2} -1).(e^{~2\mu + \sigma^2})$. We now compute the mean of the random variable $e^{\plogitval{x}{\tlabel}}$ as follows:
%
\begin{align}
  \nonumber M = \mathbb{E}\left[ \frac{p}{\pi_v(1-p)t}+ e^{\logitval{x}{\tlabel}} \left( 1+\frac{p}{\pi_v(1-p)t}\right) \right]
  \\\nonumber=  \frac{p}{\pi_v(1-p)t} + \mathbb{E}[e^{\logitval{x}{\tlabel}}] . \left( 1+\frac{p}{\pi_v(1-p)t}\right)
  \\= \frac{p}{\pi_v(1-p)t} + e^{~\mu + \sigma^2/2} .  \left( 1+\frac{p}{\pi_v(1-p)t}\right)
\end{align}
Similarly, we compute the variance of $e^{\plogitval{x}{\tlabel}}$ as
%
\begin{align}
    \nonumber V 
    = \mathbf{Var}\left[\frac{p}{\pi_v(1-p)t}
    + e^{\logitval{x}{\tlabel}} \left( 1+\frac{p}{\pi_v(1-p)t} \right) \right]
    \\\nonumber = \mathbf{Var}(e^{\logitval{x}{\tlabel}}).\left( 1+\frac{p}{\pi_v(1-p)t}\right)^2
    \\= (e^{\sigma^2} -1).(e^{~2\mu + \sigma^2}).\left( 1+\frac{p}{\pi_v(1-p)t}\right)^2
\end{align}
 For simplicity of analysis, we assume  $\plogitval{x}{\tlabel}$ also follows Gaussian  distribution with $\Tilde{u}$ and $\Tilde{\sigma}^2$ denoting its mean and variance respectively. As a result random variable $e^{\plogitval{x}{\tlabel}}$ follows a log-normal distribution with mean and variance as $M$ and $V$ respectively. We can now write a system of equations from standard log-normal definition as: $e^{\Tilde{u} + \Tilde{\sigma}^2/2} = M$ and $(e^{\Tilde{\sigma}^2-1}).(e^{2\Tilde{\mu} + \Tilde{\sigma}^2}) = V$. On solving for $\Tilde{u}$ and $\Tilde{\sigma}^2$, we get:  
\begin{align*}
    \Tilde{u} = {\log{M} - \log{\left(\sqrt{\frac{V}{M^2}+1}\right)}}
    \\  \Tilde{\sigma}^2 = \log{\left( \frac{V}{M^2} + 1\right)}
\end{align*}

\end{proof}

Note that, based on Eqn. (\ref{eqn:logit_eqn}), when the term $\frac{p}{\pi_v(1-p)t}>0$, then random variable $\plogitval{x}{\tlabel}$ does not follow a Gaussian distribution naturally. However, in practice the poisoning fraction $p$ is chosen to be a very small value and as a result  assuming $\plogitval{x}{\tlabel}$ to be Gaussian is a fair approximation.

\subsection{Computing Optimal Threshold} \label{proof:threshold}

\OptimalThreshold*

\begin{proof}
Our goal is to find threshold $\threshold$ that minimizes the objective function $J =  \alpha+\beta$. We re-write the objective function as follows:
\begin{align*}
	J = \alpha+\beta =   \Pr[X_0> \threshold] + \Pr[X_1<\threshold] 
	\\= \Pr[X_1<\threshold] + 1 - \Pr[X_0<\threshold] 
	\\= 1 + \Phi\left(\frac{\threshold-\mu_1}{\sigma_1}\right) - \Phi\left(\frac{\threshold-\mu_0}{\sigma_0}\right)
\end{align*}
where $\Phi\left(\frac{\threshold-\mu_i}{\sigma_i}\right)$ denotes the CDF of the random variable $X_i$. To compute the optimal value of $\threshold$, we differentiate $J$ with respect to  $\threshold$ and solve the equation is as follows:

\begin{align*}
	\frac{\partial J}{\partial \threshold} = \frac{\partial }{\partial \threshold}  \Phi\left(\frac{\threshold-\mu_1}{\sigma_1}\right) -  \frac{\partial}{\partial \threshold}  \Phi\left(\frac{\threshold-\mu_0}{\sigma_0}\right)  
	\\=  \frac{1}{\sigma_1}\phi\left(\frac{\threshold-\mu_1}{\sigma_1}\right) - \frac{1}{\sigma_0}\phi\left(\frac{\threshold-\mu_0}{\sigma_0}\right)
\end{align*}
where $\phi\left(\frac{\threshold-\mu_i}{\sigma_i}\right)$ denotes the PDF of the random variable $X_i$. Setting  the above equation to $0$ and substituting the gaussian PDF equation for $\phi$, we get:

\begin{align*}
	e^{{-(\threshold-\mu_1)^2}/{2\sigma_1^2}}  = \left({\sigma_1}/{\sigma_0}\right)^2  e^{{-(\threshold-\mu_0)^2}/{2\sigma_0^2}} 
\end{align*}

On re-arranging the above equation:
\begin{flalign} \label{eq:logit_val}
\nonumber (\sigma_1^2 - \sigma_0^2)\threshold^2 + 2(\mu_1\sigma_0^2 - \mu_0\sigma_1^2) \threshold 
\\+\mu_0^2\sigma_1^2 -\mu_1^2\sigma_0^2 
+ 4\sigma_1^2\sigma_0^2\log\left({\sigma_0}/{\sigma_1}\right) = 0
\end{flalign}
The roots of equation \ref{eq:logit_val} can then be written as:
\begin{equation*}
	\small
	\resizebox{.47\textwidth}{!}{$T = \frac{(\mu_0\sigma_1^2 - \mu_1\sigma_0^2) 
			\pm 2\sigma_1 \sigma_0 \sqrt{\left(\frac{\mu_1-\mu_0}{2}\right)^2 
				+(\sigma_0^2 - \sigma_1^2) \log \left(\frac{\sigma_0}{\sigma_1}\right) }}
		{\sigma_1^2 - \sigma_0^2}$}
\end{equation*}
\end{proof}  

When the standard deviations for the two Gaussians are the same, i.e. $\sigma_0 = \sigma_1$,  Eqn. (\ref{eq:logit_val}) becomes:
\begin{align*}
	 \threshold = \frac{ \mu_1^2\sigma_1^2 -\mu_0^2\sigma_1^2} {2(\mu_1\sigma_1^2 - \mu_0\sigma_1^2)} = \frac{\mu_0+ \mu_1}{2} 
\end{align*}
 
 
 \subsection{Number of Test Queries} \label{proof:nqueries}
 \TestQueries*
 
 \begin{proof}
	We  define a Bernoulli random variable $b=1$ with probability $\alpha$, and 0 otherwise,
	where $\alpha$ denotes  the probability of making Type-I errors. We predict fraction $\world{1}$  iff $\Pr[(X_0 = \sum_{i=1}^{q_0} b_{i})  > q_0/2] < \epsilon$, for some very small probability $\epsilon$. 
	
	We can then bound this probability by applying Chernoff bound, with $\delta>0$, as follows:
	\begin{align*}
		\Pr[X_0 > \frac{q_0}{2}] = \Pr[X_0 > (1+ \delta_\alpha)\mu] < e^{\frac{-\delta_\alpha^2 \mu}{2+\delta_\alpha}}
	\end{align*}
	where $\mu = \alpha \cdot q_0$ is the mean of $X_0$ and $\delta_\alpha = {1}/{2\alpha}-1$. Condition $\delta_\alpha>0$ implies $\alpha<1/2$. Now solving for $ e^{\frac{-\delta_\alpha^2 \mu}{2+\delta_\alpha}} = \epsilon$, we get $$q_0 = \frac{-(2+\delta_\alpha) \log \epsilon }{\alpha \cdot \delta_\alpha^2 }$$
	
	Similarly, we define another Bernoulli random variable $b'=1$ with probability $\beta$,
    where $\beta$ denotes the probability of making Type-II errors and we predict $\world{0}$ iff $\Pr[(X_1 = \sum_{i=1}^{q_1} b'_{i})  > q_1/2] < \epsilon$. On applying Chernoff bound and solving for $q_1$, we get
	$$q_1= \frac{-(2+\delta_\beta) \log \epsilon }{\beta \cdot \delta_\beta^2}$$
	where $\delta_\beta = \frac{1}{2\beta}-1$ and $\beta < 1/2$.  We then set the number of queries $|D_q| = \text{max}(q_0, q_1)$. 
	
\end{proof}

\section{Additional Experiments} \label{sec:AddExps}

In this section, we present additional experiments and comparison to prior work.

\subsection{Comparison with prior work \cite{Chase21}} \label{apndx:CompMSR}

\begin{figure*}[th]{
		\centering
			\includegraphics[width=0.47\textwidth]{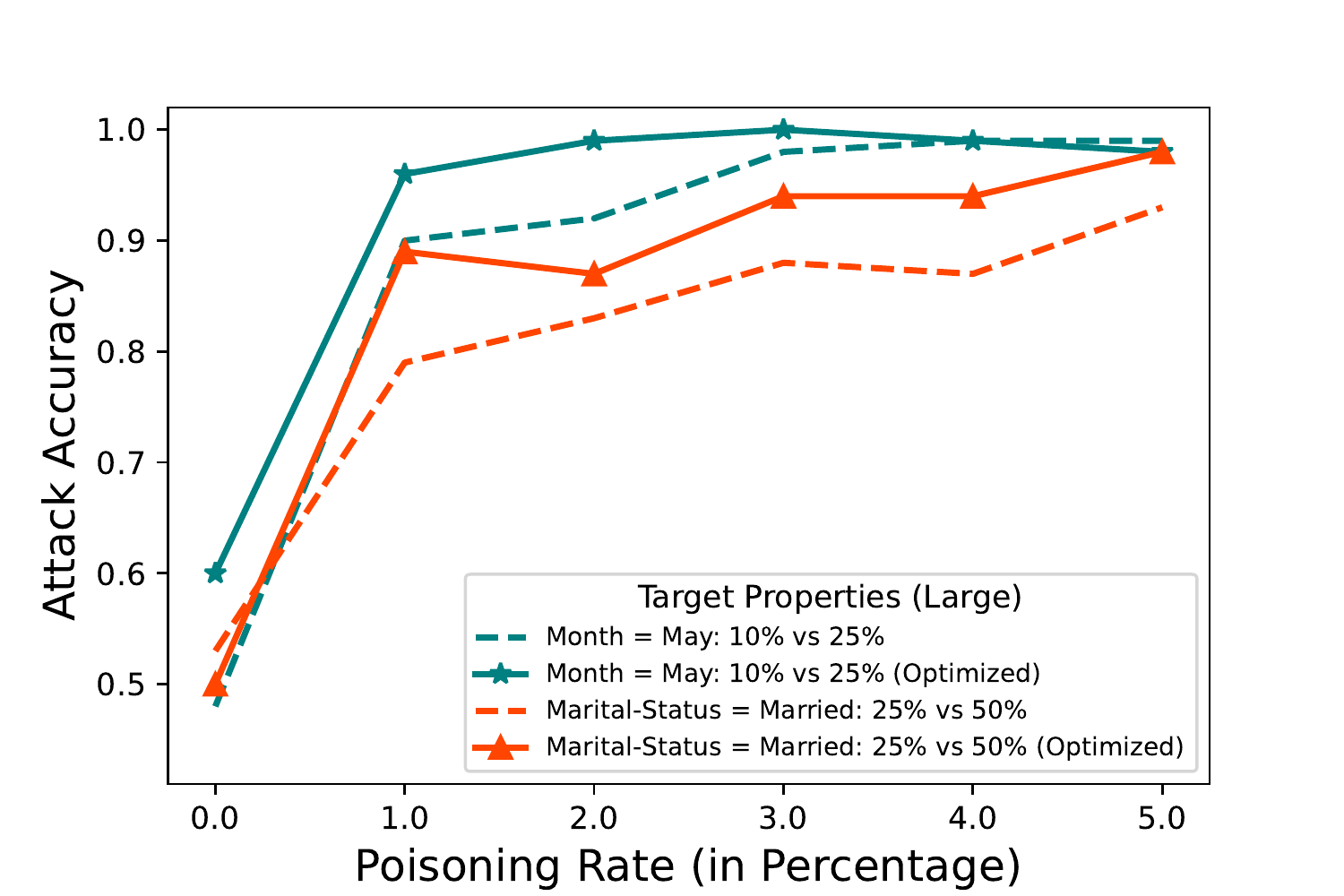}%
			\includegraphics[width=0.47\textwidth]{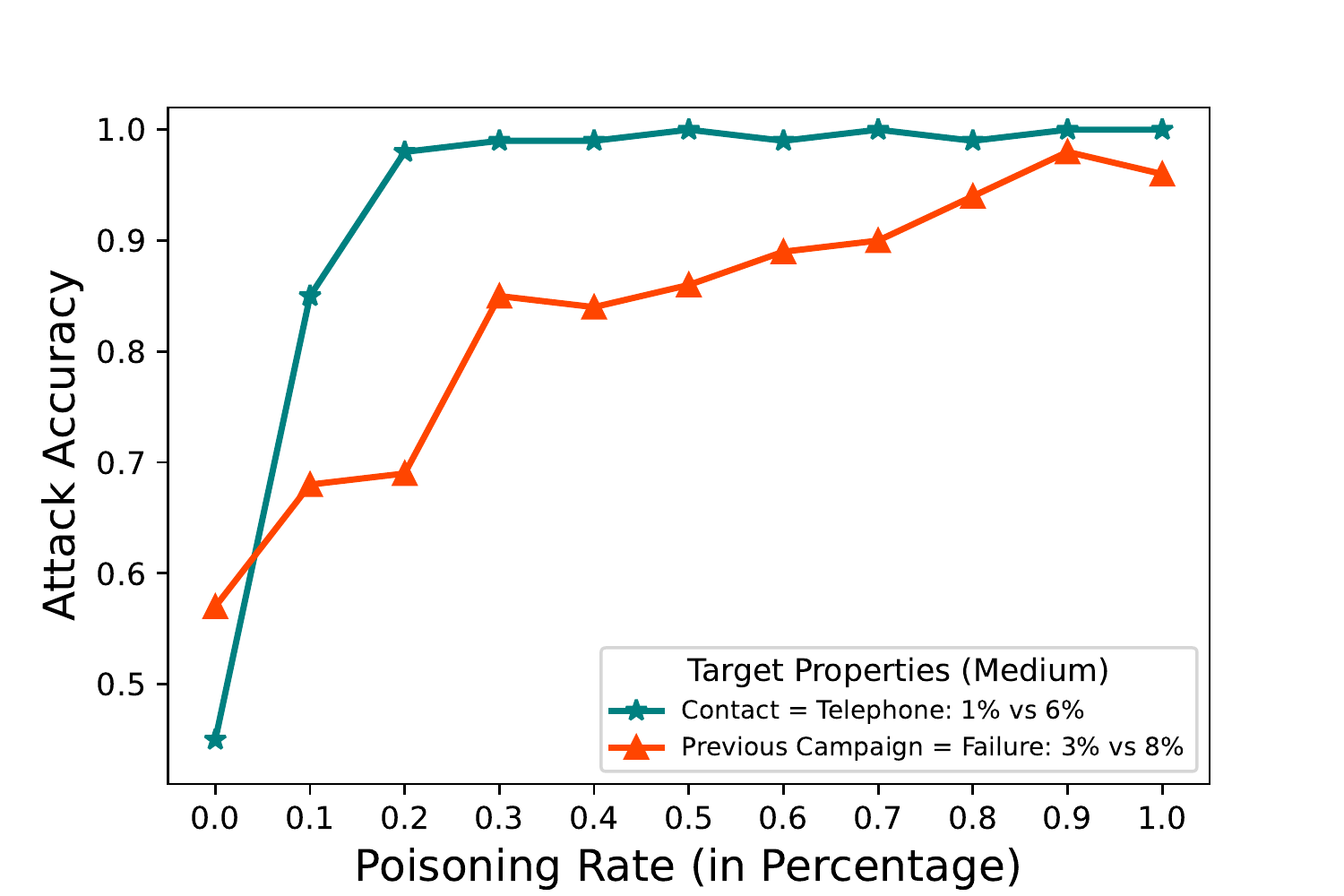}%
		\caption{Attack accuracy by poisoning rate for large and medium  properties on Bank Marketing dataset. Attack accuracy reaches $90\%$ at  0.6\% poisoning rate for medium properties. Our optimized attack for large properties consistently outperforms our original attack.}
    	\label{fig:apndx_BM}
}
\end{figure*}

We  compare our attack strategy to~\cite{Chase21} on the CelebA dataset, on the target property ``Gender = Male'' and the percentages as $30\%$ or $70\%$ on a smile detection task. We report the results in Table~\ref{tab:CompCelebA}. We use two ResNet-18 shadow models per fraction and train each model for 30 epochs using Adam with a learning rate of $0.03$ and a batch size of 64. Before we run our attack, we ensure that our target models have achieved high accuracy, precision, and recall. We run the attack for 5 trials, each containing 20 queries to the target model. On average, each trial takes 1 hour and 40 minutes on 32 Intel Xeon E5-2680 CPU threads and one Nvidia Titan X (Pascal) GPU.

{\begin{table}[h!]
		\centering \scriptsize
		\begin{adjustbox}{max width=0.7\textwidth}{  
				\begin{tabular}{c c c c c c}
					
					
					 \multirow{2}{*}{} & \multirow{2}{*}{Attack Strategy}  & \multirow{2}{*}{$\#$ Shadow Models} & \multicolumn{3}{c}{Poisoning Rate} \\ 
					
					\cmidrule{4-6}
					
					                           &                                   &                                     & \multicolumn{1}{c}{$0\%$} & \multicolumn{1}{c}{ $5\%$} & \multicolumn{1}{c}{$10\%$} \\
					\midrule
					
					                          & Mahloujifar et al. \cite{Chase21} & $500$ & $\mathbf{73\%}$  & $92\%$ & $97\%$\\
					                          & \system\ (Ours)         & $\mathbf{2}$ & $47\%$  & $\mathbf{100\%}$ & $\mathbf{100\%}$\\                         
					 

				\end{tabular}
			}
		\end{adjustbox}
		\caption{Attack accuracy comparison with \cite{Chase21} using ResNet-18 as the model architecture trained on CelebA dataset with the target property ``Gender = Male.''} \label{tab:CompCelebA}
	\end{table}
}

\subsection{\revision{Label-Only Evaluation}} 
\label{apndx:Label-Only}
\begin{figure}[th!]
 \vspace{-0.5cm}
 \centering
	\includegraphics[width=0.5\textwidth]{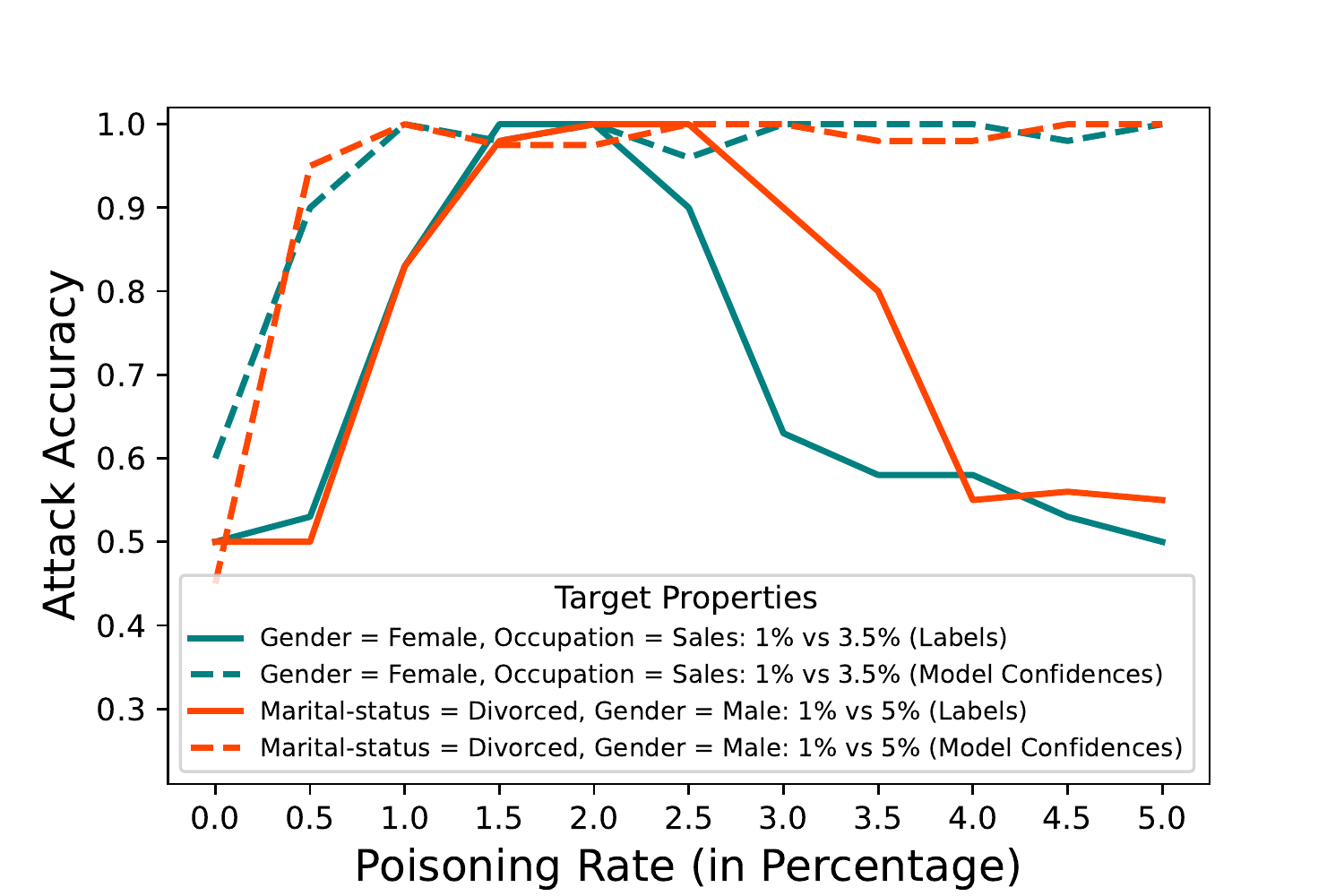}%
	\caption{\revision{Behavior of Label-only and Model confidence  versions of  \system. Attack accuracy  for the confidence version increases with the amount of poisoning, while the attack accuracy increases and then drops  for Label-only  version, making it crucial to choose an appropriate  poisoning rate.}}
	\label{fig:CompLabelOnly}
 \vspace{-2mm}
\end{figure}

\revision{Recall that in our label-only extension, choosing an appropriate poisoning rate $p^*$ is crucial for our attack to succeed. Figure~\ref{fig:CompLabelOnly} shows one such instance where the attack accuracy for the label-only \system\ attack  is as high as our model confidence version  for a small range of poisoning rates. Our approach of computing a suitable poisoning rate $p^*$ indeed gives us a high attack success on the distinguishing task. For instance, given the target property  `Female Sales' in Figure \ref{fig:CompLabelOnly}, our approach suggests a poisoning rate $p^*$ of $1.23\%$, for which we achieve an attack accuracy of $98\%$.}

\subsection{Additional Properties}

We perform experiments on the remaining properties considered for Bank Marketing and CelebA datasets. Table~\ref{tab:apndxTP} summarizes the target properties associated to these datasets.

{\begin{table}[h!]
		\centering 
		\begin{adjustbox}{max width=0.7\textwidth}{  
				\begin{tabular}{c c c  r  r}
					
					
					{Attack Type} & {Property Size} & {Dataset} & \multicolumn{1}{c}{Target Properties} & Distinguishing Test\\ \midrule
					
					\multirow{4}{*}{\makecell{Property \\Inference}} & \multirow{2}{*}{Large}  & \multirow{2}{*}{Bank Marketing} & Month = May & $10\%$ vs $25\%$ \\
					
					                                    &                         &                        &  Marital-Status = Married & $25\%$ vs $50\%$ \\
                    
                  \cmidrule{3-5}
                    
                                                     &                         & \multirow{3}{*}{CelebA} & Gender = Male & $30\%$ vs $70\%$\\     
                                                     &                         &                          & Age = Old & $25\%$ vs $60\%$\\    
                                                     &                         &                          & Wearing Earrings & $15\%$ vs $40\%$\\

                    \cmidrule{2-5}
                                                        & \multirow{2}{*}{Medium} & \multirow{2}{*}{Bank Marketing}   & Contact = Telephone & $1\%$ vs $6\%$ \\
                                                        
                                                        &                        &                          &  Previous Campaign = Failure & $3\%$ vs $8\%$ \\ 
                                                        

				\end{tabular}
			}
		\end{adjustbox}
		\caption{Properties considered in Bank Marketing and CelebA datasets. The attacker's objective is to distinguish between the two percentages  of the target property. } 
		\label{tab:apndxTP}
	\end{table}
}

Figure~\ref{fig:apndx_BM} provides results of attack accuracy for large and medium target properties on the Bank Marketing dataset. We observe the attack accuracy improves dramatically across all properties as the poisoning rate increases. The attack accuracy reaches $90\%$ with as little as 0.6\% poisoning for medium properties. For large properties, we observe the optimized variant consistently outperforms our original variant, obtaining close to $90\%$  accuracy with only 1\% poisoning.  

We run \system\ on two more properties from CelebA: Older Faces (Young = 0) and Wearing Earrings with the classification tasks of smile prediction and gender prediction, respectively. The Young property has been used in prior work on property inference \cite{Ganju18, DistributionInference}. Our distinguishing tests for these two properties are 25\% vs 60\%  and 15\% vs 40\%, respectively. For 0\%, 5\%, and 10\% poisoning on Older Faces, our attack success was 57\%, 99\%, and 99\%. For 0\%, 5\%, and 10\% poisoning on Wearing Earrings, our attack success was 40\%, 73\%, and 79\%. This shows that \system\ works on computer vision tasks, which are in general challenging for meta classifier-based property inference attacks.

\myparagraph{Sub-properties used for our optimized attack}
Table~\ref{tab:apndxSubP} provides the  sub-properties used in our optimized \system\ attack, when targeting large properties for the Adult, Census and Bank Marketing datasets. Recall that the adversary poisons the sub-property within the large property to distinguish between the two fractions  of the large property as given in the last column of Table \ref{tab:apndxSubP}.

{\begin{table}[h!]
		\centering 
		\begin{adjustbox}{max width=0.7\textwidth}{  
				\begin{tabular}{c r  r  r}
					
					
					 {Dataset} & \multicolumn{1}{c}{Target Properties} & \multicolumn{1}{c}{Sub-Properties} & Distinguishing Test\\ \midrule
					
					\multirow{2}{*}{Adult} & Workclass = Private & Occupation = Transportation & $20\%$ vs $40\%$ \\
					
                     & Race = White; Gender = Male &  Marital-Status = Never-Married & $15\%$ vs $30\%$ \\
                    
                    \cmidrule{1-4}
                      \multirow{2}{*}{Census} & Race = Black & Education = High-School & $10\%$ vs $25\%$ \\
					
                                               & Gender = Female &  Race = Black & $30\%$ vs $50\%$ \\
                    
                    \cmidrule{1-4}
                        \multirow{2}{*}{Bank Marketing} & Month = May & Occupation = Technician & $10\%$ vs $25\%$ \\
					
                                              & Marital-Status = Married &  Month = July & $25\%$ vs $50\%$ \\

				\end{tabular}
			}
		\end{adjustbox}
		\caption{Sub-properties considered for our optimized attack on Adult, Census and Bank Marketing datasets. } 
		\label{tab:apndxSubP}
		\vspace{-0.5cm}
	\end{table}
}

\subsection{\revision{Experiments with Differential Privacy}} \label{apndx:DP}
\revision{The target and shadow model architectures for our experiments on Census and Adult were 4NN and 3NN, respectively. We used PyTorch's differential privacy library, Opacus \cite{opacus}, to train all of the neural network models. Each model was trained using the Adam optimizer \cite{AdamOptimizer} with the Opacus wrapper for 40 epochs with a batch size of 512, a learning rate of 0.003, and a clipping threshold of 1.2. The poisoning rate was set to 4\% for Census and 6\% for Adult. The privacy parameters $(\varepsilon, \delta)$ were chosen ahead of time, and Opacus tracked the remaining privacy budget at each training epoch.
}

\begin{table}[h!]
\centering
\resizebox{0.65\columnwidth}{!}{%
\begin{tabular}{rccccc}

\multicolumn{1}{c}{\multirow{3}{*}{Target Property}} &  & \multicolumn{4}{c}{Attack Success $(\varepsilon, \delta=10^{-5})$} \\ \cmidrule{3-6}
             &  & \multicolumn{1}{c}{$\varepsilon=8$} & \multicolumn{1}{c}{$\varepsilon=4$} & \multicolumn{1}{c}{$\varepsilon=2$} & \multicolumn{1}{c}{$\varepsilon=1$} \\ \midrule
Race = Black                &   & 100\%  & 100\% & 100\%  & 98\%  \\ 
Race = White, Gender = Male  &   & 95\%  & 99\% & 90\% & 75\% \\
\end{tabular}%
}
\caption{Attack success of \system\ when the target model is trained using DP-Adam for several privacy parameters.}
\label{tab:DPAdam}
\end{table}

The success of \system\ begins to decrease once private training decreases the utility of the model. For instance, the attack success is lowered to 75\% on the ``Race = White; Gender = Male'' property for $\epsilon =1$, but the F1 score of the private target model on the subpopulation is only 0.07.

\end{document}